\pdfoutput=1
\documentclass[10pt,twocolumn,letterpaper]{article}

\usepackage[pagenumbers]{cvpr} % To force page numbers, e.g. for an arXiv version

\usepackage{graphicx}
\usepackage{amsmath}
\usepackage{amssymb}
\usepackage{booktabs}
\usepackage{xcolor}
\usepackage{algorithm} 
\usepackage{algpseudocode} 

\usepackage{enumitem}
\usepackage[accsupp]{axessibility}
\setlist{nosep}

\usepackage[pagebackref,breaklinks,colorlinks]{hyperref}

\usepackage[capitalize]{cleveref}
\crefname{section}{Sec.}{Secs.}
\Crefname{section}{Section}{Sections}
\Crefname{table}{Table}{Tables}
\crefname{table}{Tab.}{Tabs.}

\pdfoutput=1
\def\cvprPaperID{9480} % *** Enter the CVPR Paper ID here
\def\confName{CVPR}
\def\confYear{2022}

\title{SIMBAR: Single Image-Based Scene Relighting For Effective Data Augmentation For Automated Driving Vision Tasks}

\author{Xianling Zhang\textsuperscript{1*}, 
Nathan Tseng\textsuperscript{2*}, 
Ameerah Syed\textsuperscript{1}, 
Rohan Bhasin\textsuperscript{1},
Nikita Jaipuria\textsuperscript{1} \\
\textsuperscript{1}Ford Greenfield Labs, Palo Alto \qquad \textsuperscript{2}University of Michigan\\
\tt \footnotesize \{xzhan258, asyed17, rbhasin, njaipuri\}@ford.com, tsnathan@umich.edu
}

\begin{document}
\maketitle

%%%%%%%%% ABSTRACT
\begin{abstract}
Real-world autonomous driving datasets comprise of images aggregated from different drives on the road. The ability to relight captured scenes to unseen lighting conditions, in a controllable manner, presents an opportunity to augment datasets with a richer variety of lighting conditions, similar to what would be encountered in the real-world. This paper presents a novel image-based relighting pipeline, SIMBAR, that can work with a single image as input. To the best of our knowledge, there is no prior work on scene relighting leveraging explicit geometric representations from a single image. We present qualitative comparisons with prior multi-view scene relighting baselines. To further validate and effectively quantify the benefit of leveraging SIMBAR for data augmentation for automated driving vision tasks, object detection and tracking experiments are conducted with a state-of-the-art method, a Multiple Object Tracking Accuracy (MOTA) of 93.3\% is achieved with CenterTrack on SIMBAR-augmented KITTI - an impressive 9.0\% relative improvement over the baseline MOTA of 85.6\% with CenterTrack on original KITTI, both models trained from scratch and tested on Virtual KITTI. For more details and relit datasets, please visit our project website (\url{https://simbarv1.github.io}). 
\end{abstract}
\vspace{-0.05in}

%%%%%%%%% BODY TEXT
\vspace{-.2in}
\section{Introduction}
\label{sec:intro}
A lack of diversity in lighting conditions is a known issue with manually collected real-world autonomous driving datasets \cite{DBLP:journals/corr/Aly14, Agustsson_2017_CVPR_Workshops, Fritsch2013ITSC, geiger2013vision}. For example, KITTI \cite{geiger2013vision} has video sequences captured only during noon, with similar lighting and shadow conditions across different sequences.
% during daytime, and that too at very similar times each day. 
More recent datasets\cite{yu2018bdd100k, Lee_2017_ICCV, MVD2017}, one such as BDD100K \cite{yu2018bdd100k}, are comparatively better in terms of diversity and have images captured during multiple times of the day. Still, between images collected from the same drive, there are minimal changes in lighting conditions. Furthermore, attempting to acquire data for all types of lighting conditions is implausible both in terms of time and money.

\begin{figure}[ht]
\centering
\includegraphics[width=0.47\textwidth]{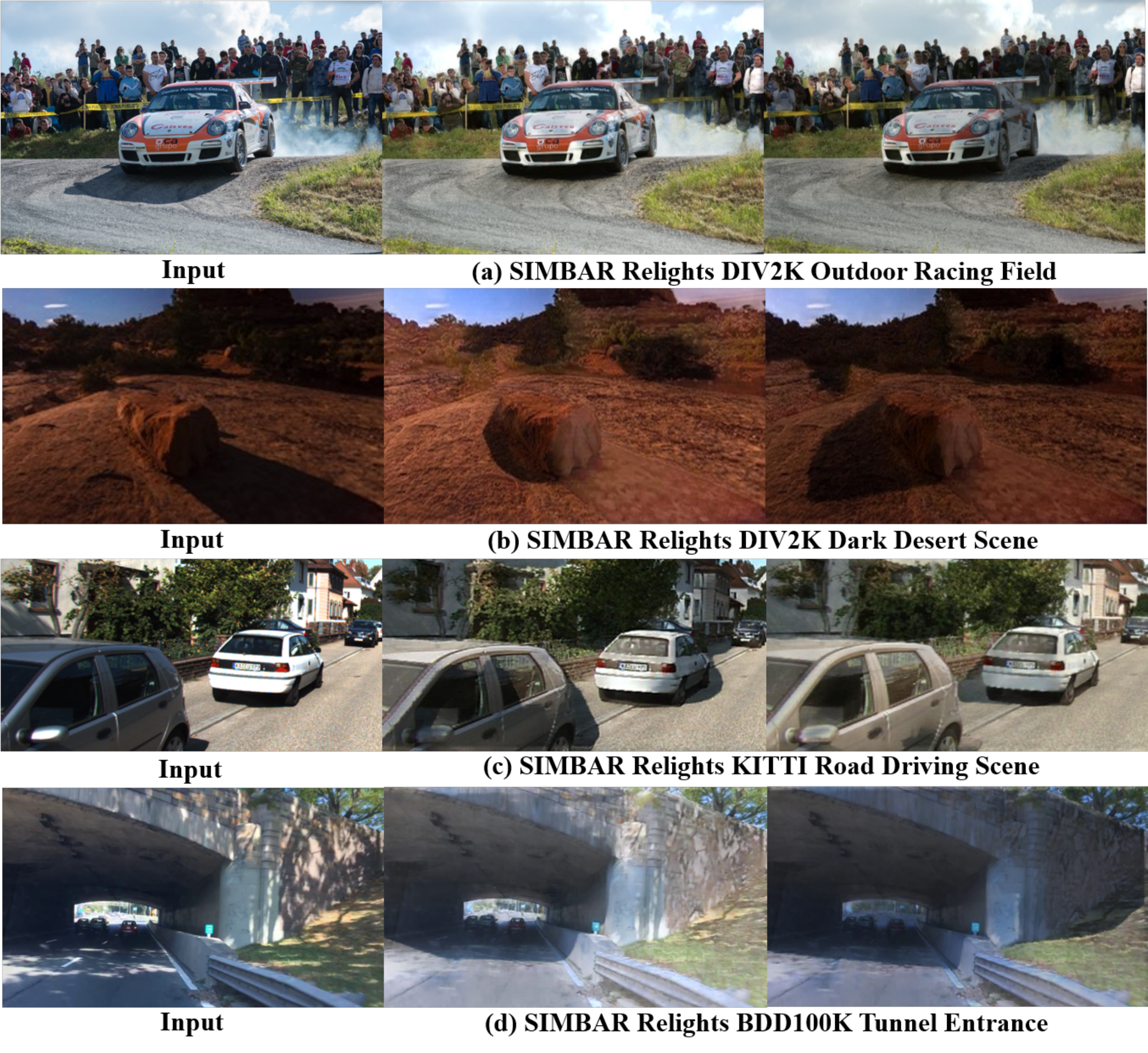}
\vspace{-.1in}
\caption{\small Input images (Left) are shown against SIMBAR-relit outputs (Middle, Right). SIMBAR synthesized two lighting variations for (a)(b) Div2k, (c) BDD100K and (d) KITTI. 
\label{fig:intro_sr_relit_example}}
\end{figure}
% \vspace{-.2in}

This lack of diversity in lighting conditions and by extension, the shadows present within a scene, often serves as a crucial roadblock in successful real-world deployment of perception models for safety-critical automated driving applications. Models trained with limited lighting conditions are unable to generalize to the plethora of lighting conditions encountered in the real-world \cite{ibarra2020shadow, jaipuria2020deflating}. The ability to relight existing datasets in a controllable manner presents an opportunity to develop improved perception models.
However, scene relighting, in the absence of depth sensors, is an extremely difficult vision task. It implicitly comprises of three main sub-tasks: shadow detection\cite{DBLP:conf/aaai/CunPS20,DBLP:conf/cvpr/Hu0F0H18, DBLP:conf/cvpr/WangH0HF20}, removal\cite{DBLP:journals/pami/HuFZQH20, DBLP:conf/iccv/HuJFH19, DBLP:conf/cvpr/WangH0HF20 } and insertion\cite{DBLP:journals/cvm/ZhangLW19}. Of these, shadow removal and insertion are most challenging because shadows blend tightly with the source object geometry \cite{al2012survey, fu2020learning}. This coupling makes it difficult to separate the shadow from its parent object without a strong 3D geometric understanding of the scene\cite{DBLP:journals/corr/abs-1806-01260, DBLP:journals/corr/abs-1709-06158,DBLP:journals/corr/ArmeniSZS17}. To address this, most prior scene relighting methods rely on multiple camera views of the source lighting condition to estimate the 3D scene geometry \cite{srinivasan2021nerv,zhang2021nerfactor,philip2019multi}. The relatively few prior methods that can work with a single image are based on Generative Adversarial Networks (GANs) \cite{carlson2019shadow}. GANs are known to be difficult to train \cite{kodali2017convergence, neyshabur2017stabilizing}, limited in controllability \cite{voynov2020unsupervised}, and often produce results that are physically inconsistent with scene geometry \cite{gafton20202d}. To the best of our knowledge, there is no prior work on \emph{controllable} scene relighting using a single input image.

This paper presents a novel \textbf{S}ingle \textbf{IM}age-\textbf{BA}sed scene \textbf{R}elighting pipeline, SIMBAR. It takes a single image as input and produces relit versions for a wide variety of sun positions and sky zeniths, as shown in Fig.~\ref{fig:intro_sr_relit_example}. The top two rows show relit results from the Div2k \cite{Agustsson_2017_CVPR_Workshops}. Div2k is an internet-scraped dataset with images of a wide variety of object classes, that SIMBAR is able to effectively relight. The first row shows realistic variations in sky colors, shadow orientations, and consistent cast shadow locations and light intensities for an outdoor scene with complex structures. The second row is a challenging low-light desert scene. SIMBAR cleanly removes existing hard cast shadows of the rock in the foreground and realistically recasts geometrically consistent shadows for the provided sun angle. Additionally, the mountainous landscape in the horizon has also been effectively relit. The third and fourth rows also show geometrically consistent and visually realistic relit versions of a KITTI road driving scene and a tunnel/underpass scene from BDD100K  respectively. Most notable is the variation in hard cast shadows of the tunnel in the BDD100K example and the two cars in the KITTI example.

SIMBAR consists of two main modules: (i) geometry estimation and (ii) image relighting. The geometry estimation module is responsible for computing scene mesh proxy and illumination buffers. We are inspired by WorldSheet\cite{hu2021worldsheet} to use external depth networks to obtain a scene mesh. Note that WorldSheet is a novel view synthesis pipeline that does not have relighting purpose. The image relighting module is inspired by prior work on multi-view scene relighting using a geometry aware network \cite{philip2019multi}, referred to as MVR for brevity. Section~\ref{subsec:prelim} provides a short overview of Single Image-Based Scene Geometry Estimation and MVR, followed by a detailed description of SIMBAR's pipeline description in Section~\ref{subsec:method}. Our work is closest in terms of goals and overall pipeline structure to MVR. Therefore, scene relighting comparisons are provided with both out-of-the-box MVR and its improved version, MVR-I, where we refined MVR for autonomous driving datasets with limited views, in Section~\ref{subsec:relit_results}. Across the board, SIMBAR provides significantly more realistic and geometrically consistent relit images, even though it takes as input a single image, as compared to MVR/MVR-I that take as input multiple images of the same scene.

Another major limitation of all prior works on scene relighting is the lack of a quantitative evaluation of the effectiveness of scene relighting in augmenting vision datasets. In the absence of such a metric, the real-world applicability and usefulness of any scene relighting methodology cannot be established. To address this, in Section~\ref{sec:augmentation}, we perform image relighting-based data augmentation experiments with a state-of-the-art object detection and tracking network, CenterTrack \cite{zhou2020tracking}. Section~\ref{subsec:experiment_setup} provides a detailed overview of our experiment setup. We train three different CenterTrack models on: (i) original KITTI tracking dataset with 21 real-world sequences captured at noon; (ii) augmented KITTI with MVR-I relit sequences; and (iii) augmented KITTI with SIMBAR relit sequences. All models are tested on Virtual KITTI (vKITTI) \cite{gaidon2016virtual}, which consists of clones of real KITTI sequences in a variety of lighting conditions. Section~\ref{subsec:quant_comparison} shows that CenterTrack models augmented with relit KITTI images (from either MVR-I or SIMBAR) consistently outperform the baseline CenterTrack model. Specifically, the CenterTrack model trained on KITTI augmented with SIMBAR achieves the highest Multiple Object Tracking Accuracy (MOTA) of 93.3\% - a 9.0\% relative improvement over the baseline MOTA of 85.6\%. This model also achieves the highest Multiple Object Detection Accuracy (MODA) of 94.1\% - again an impressive 8.9\% relative improvement over the baseline MODA of 86.4\%.

To summarize, the main contributions of this paper are:
\begin{enumerate}
    \item A novel single-view image-based scene relighting pipeline, called SIMBAR, that offers lighting controllability without the need for multi-perspective images. 
    \item Single image-based geometry estimation via adapting dense prediction transformer monodepth model and better representation of far-away background objects.
    \item An improved version of MVR\cite{philip2019multi}, called MVR-I, with fewer artifacts and smoother surfaces in the generated mesh for road driving scenes with limited views, resulting in more realistic relit images.
    \item Qualitative evaluation and comparison of scene relighting results using MVR, MVR-I and SIMBAR, on multiple automated driving datasets, such as KITTI\cite{geiger2013vision} and BDD100K\cite{yu2018bdd100k}.
    % \item Quantitative evaluation and comparison of the real-world application effectiveness of augmenting datasets with data from SIMBAR and MVR-I, using CenterTrack \cite{zhou2020tracking} and KITTI dataset \cite{geiger2013vision}.
    \item Quantitative evaluation of the effectiveness of augmenting the popular KITTI 2D tracking dataset using SIMBAR and MVR-I for simultaneous object detection and tracking using CenterTrack.
\end{enumerate}
% \vspace{-.05in}
% \caption{\small 3 representative lighting and shadow conditions from SIMBAR relit outputs (middle and right), from Div2k \cite{Agustsson_2017_CVPR_Workshops} (top two rows), BDD100K \cite{yu2018bdd100k} (third row) and KITTI \cite{geiger2013vision} (last row).
% \begin{figure}[ht]
% \centering
    % \begin{subfigure}[b]{0.5\textwidth}
    % \includegraphics[width=1\columnwidth]{latex/figures/section1_intro/intro_wild_simbar_div2k_c3_noted.png}\hspace{1em}%
    % \caption*{
    % \label{fig:1a}}
    % \end{subfigure}
    % \begin{subfigure}[b]{0.5\textwidth}

    % \includegraphics[width=1\columnwidth]{latex/figures/section1_intro/intro_wild_simbar_bdd_c3_noted.png}\hspace{-1em}%
    % \caption*{
    % \label{fig:1b}}
    % \end{subfigure}
% \caption{\small SIMBAR generated lighting variations for images from Div2k (a) and (b), BDD100K (c) and KITTI (d).
% \label{fig:intro_sr_relit_example}}
% \vspace{-.2in}
% \end{figure}
\section{Related Work}
\label{sec:related}
\vspace{-.05in}
Our work is closely related to the fields of novel view synthesis\cite{DBLP:journals/corr/abs-2004-04727,DBLP:journals/corr/abs-1912-08804,DBLP:journals/corr/abs-1905-00889}, 3D reconstruction\cite{DBLP:conf/cvpr/00060WNMCS21, DBLP:conf/cvpr/ChenZ19, DBLP:conf/nips/XuWCMN19}, and physics-based differentiable rendering\cite{DBLP:conf/cvpr/0002JHZ20, DBLP:conf/cvpr/ParkMXF20}. Given the direct connection between the relighting task and scene geometry\cite{DBLP:conf/cvpr/ZengSNFXF17,DBLP:conf/siggraph/DebevecTM96,DBLP:journals/tog/ZitnickKUWS04}, we split the related work into two broad categories: 
(i) implicit approaches learning geometric priors and encoding them into a model; and (ii) explicit approaches leveraging multiple views of the input scene to generate a 3D mesh to apply rendering and image processing techniques upon. While explicit approaches provide better controllability and geometrically consistent shadows, their multi-view prerequisite inhibits their application to most automated driving datasets. This is due to the unique challenge of limited views from a front-facing car camera, compounded by high scene complexity of ever-moving cars and pedestrians. Our work falls in the explicit category, while leveraging insights from the implicit approaches.

\subsection{Using Implicit Geometric Representations}
Both Generative Adversarial Networks (GANs) \cite{goodfellow2020generative} and Neural Radiance Fields (NeRFs) \cite{mildenhall2020nerf} have explored scene relighting. As is typical of GANs, the shadow manipulation network from \cite{carlson2019shadow} struggles to maintain geometric consistency and is difficult to train, thus resulting in conservative relighting effects. This also occurs for GANs that focus on image-to-image translation and ignore geometric priors \cite{gafton20202d, das2021msr}. The recent success of NeRF-based methods for novel view synthesis has naturally resulted in their application to the scene relighting task as well. Rather than querying an explicit scene geometry, NeRFs encode the scene into a multilayer perceptron (MLP)\cite{MultilayePerceptron}, which takes as input a viewing direction and location to output color and density values, that are then used for volumetric rendering\cite{DBLP_conf_cvpr_NiemeyerMOG20,DBLP:conf/cvpr/NiemeyerMOG20}. At training time, many different views of a static scene are given to the network to learn the scene geometry. At test time, the input viewing direction and location are used to render the scene with accurate lighting and shadows. Recent works have repurposed NeRFs for scene relighting by modeling the surface material and reflectance properties \cite{srinivasan2021nerv,zhang2021nerfactor, boss2021nerd}.
% The geometry learned by the NeRF feeds into another MLP along with additional information given by normal maps, light visibility maps, albedo maps, and Bidirectional Reflectance Distribution Functions \cite{srinivasan2021nerv,zhang2021nerfactor}.
However, such methods face a significant computational roadblock in their application to automated driving datasets with dynamic scenes, since each scene requires training a different model.
% For these relighting approaches, a variety of different information encodings serve as input and are important to provide strong signal to the network on shadow removal and synthesis. We therefore additionally investigate explicit geometric approaches, where the mesh can be used to generate different input maps encoding geometry information.

\subsection{Using Explicit Geometric Representations}
Combining Structure-from-Motion with Multi-View Stereo (SFM+MVS) is a common way of modeling scene geometry. It relies on feature matching across images captured from different views of a single scene of interest. After the application of SFM+MVS, bundle adjustment \cite{triggs1999bundle} can be used to generate a 3D point cloud, as is the case in COLMAP \cite{schoenberger2016mvs,schoenberger2016sfm}. The point cloud allows for application of traditional mesh reconstruction techniques, such as Delaunay \cite{cazals2006delaunay} or Poisson \cite{kazhdan2006poisson} reconstruction, to generate an explicit geometric representation of the scene. Vision tasks that utilize geometric priors, such as novel view synthesis, can take advantage of such an explicit scene representation \cite{riegler2020free,yoon2020novel}. The mesh can also be applied towards scene relighting tasks, as explored by \cite{philip2019multi}. In their work, physics-based rendering is used to approximate shadow locations using the generated mesh, with an additional network for shadow refinement. The relighting results are realistic and geometrically consistent. However, this method is severely limited in its application to a wide variety of datasets. For example, limited views and dynamic scenes result in failed mesh reconstruction \cite{innmann2020nrmvs}. In the case of relatively simpler and restricted datasets, such as human portraits, image relighting using a single view has been successful, owing to the high similarity in structure across facial data \cite{zhou2019deep,nestmeyer2020learning}. However, the same is not true for datasets of outdoor scenes, which contain a wider variety of structure and content \cite{einabadi2021deep}.
\vspace{-.15in}
% Furthermore, to the best of our knowledge, the effectiveness of relighting has not been explored as a direct data augmentation for downstream vision tasks when using real data, and previously has been confined strictly to synthetic data. This offers an opportunity to use single-view relighting to generate more varied training sets that account for real-world lighting conditions.  

% completely cut-out paragraph since worldsheet is view synthesis and covered in much more detail in the following section
% Another explicit geometry approach has been explored in view synthesis, with the use of a deformable mesh sheet to fit a scene geometry. The network pipeline learns end-to-end with predictions on offsets to deform the sheet and before painting the image for the target view \cite{hu2021worldsheet}. In the case of relighting, the majority of the latter half of the pipeline is focused on the view synthesis task, and we adapt the mesh sheet representation to work with a single-view relighting pipeline. The flexibility offered by single-view approaches allows relighting as a data augmentation to be leveraged widely across different datasets. To our knowledge, the effectiveness of relighting has not been explored as a direct data augmentation for downstream vision tasks when using real data, and previously has been confined strictly to synthetic data. This offers an opportunity to use single-view relighting to generate more varied training sets that account for real-world lighting conditions.  

\vspace{-.05in}
\section{Single Image-Based Scene Relighting}
\label{sec:relighting}
% \vspace{-.1in}
Our proposed pipeline, SIMBAR, models the scene as a 3D mesh to explicitly represent scene geometry. Physics-based rendering is then used in conjunction with a shadow refinement network to produce realistic shadow maps. The original image can be composited with the target shadow maps to form the final relit output. Such an approach addresses the limitations posed by prior works on multi-view scene relighting and can generalize across scenes.

\begin{figure*}[ht]
\centering
    \includegraphics[width=0.95\textwidth]{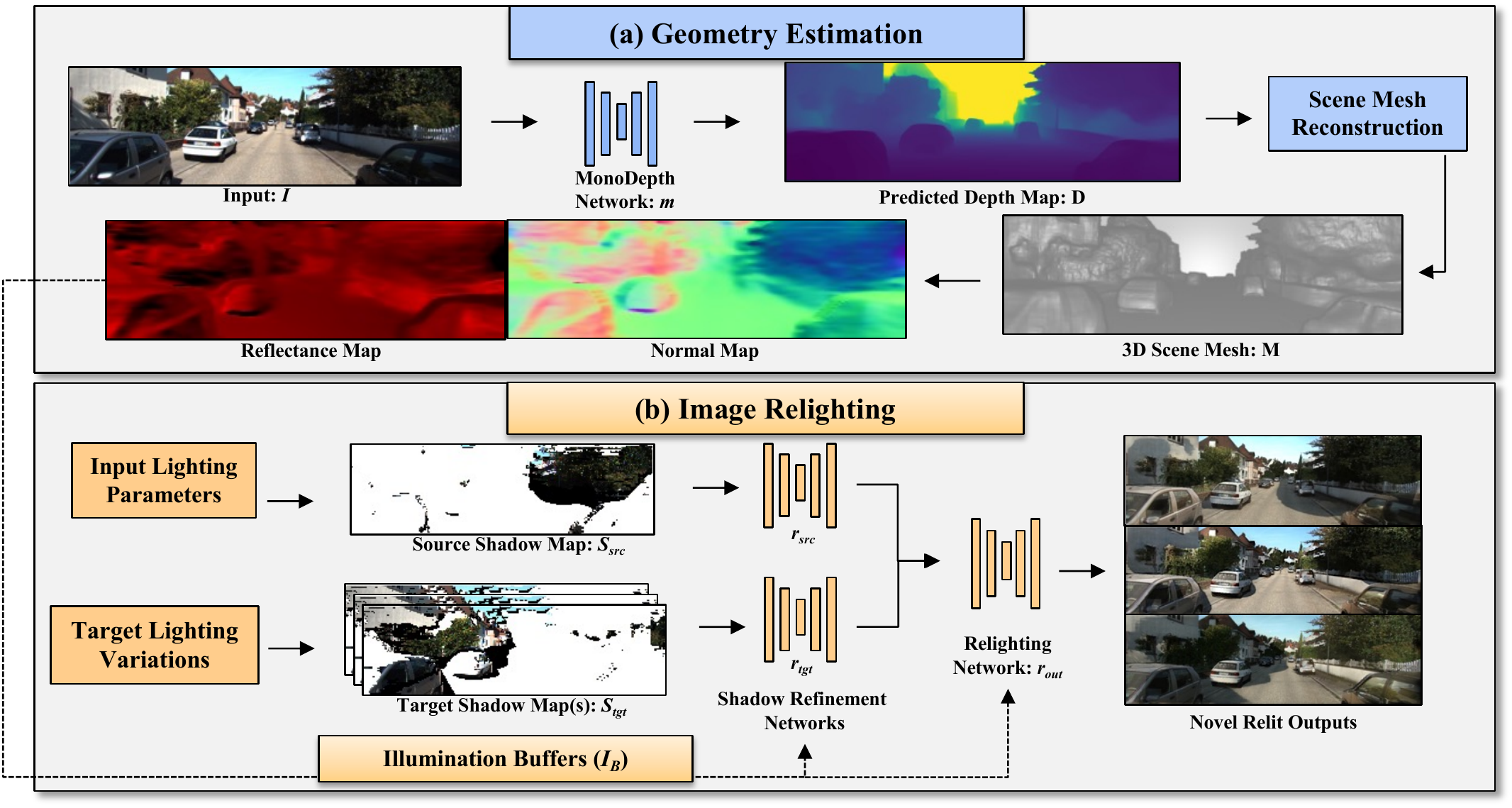}
  \vspace{-.1in}
  \caption{ \textbf{(a) Geometry Estimation Component}: a single input image, $I$, is fed to monocular depth estimation networks ($m$). The predicted depth map, $D$, is used to form scene mesh using the vertex coordinates in Eq. 1. The resulting set of vertices and faces forms the 3D mesh, $M$. A set of input buffers, $I_{B}$, are rendered with respect to the camera pose using $M$. \textbf{(b) Image Relighting Component}: With estimated input lighting parameters and demanded target lighting variations, source shadow map $S_{src}$ and target shadow map(s) $S_{tgt}$ are generated. The shadow refinement networks $r_{src}$ and $r_{tgt}$ refine the shadow maps $S_{src}$ and $S_{tgt}$ respectively. Finally, relighting network $r_{out}$ takes refined shadow maps with $I_{B}$, to generate the final relit images.
  \label{fig:simbar_pipeline}}
%   \vspace{-.1in}
\end{figure*}

\subsection{Preliminaries}
\label{subsec:prelim}
\subsubsection{Single Image-Based Scene Geometry Estimation}
\label{subsubsec:worldsheet}
% \vspace{-0.1in}
To solve the multi-view limitations of SFM+MVS-based mesh reconstruction, we have been inspired by WorldSheet\cite{hu2021worldsheet} to use external depth for scene geometry estimation in order to perform single image-based mesh reconstruction. Note that the underlying ideas for the overall WorldSheet and SIMBAR pipelines are completely different. WorldSheet is a differentiable rendering pipeline, trained end-to-end for novel view synthesis, while SIMBAR is designed to manipulate existing views with various shadows cast. 

For scene mesh formation, external depth predictions are treated as ground truth, thus requiring no predictions for grid offsets in the $x$ and $y$ direction. Let $z_{w,h}$ be the depth prediction at the corresponding sheet coordinate $(w,h)$ and $x_{w,h}$ and $y_{w,h}$ are simply linearly spaced samples in the Normalized Device Coordinates (NDC) space from $[0, 1]$, with the camera placed at the origin. Given a fixed size of the mesh sheet of $129\times 129$, depth predictions are grid-sampled to account for differences in resolution. With FoV angle, $\theta_F$, this gives the following equation for forming the vertex coordinates:
\begin{equation}
V_{w,h} = \begin{bmatrix} z_{w,h}~x_{w,h}~tan(\theta_f/2) \\ 
 z_{w,h}~y_{w,h} tan(\theta_f/2) \\
 z_{w,h}
\end{bmatrix}
\end{equation}
Grid edges that connect neighboring vertices form the mesh faces \cite{hu2021worldsheet}. The faces are then smoothed with a Laplacian function\cite{laplacian_surface_editing} for the final output mesh.
% Grid edges connect neighboring vertices to form mesh faces \cite{hu2021worldsheet}, which are smoothed with a laplacian function \cite{laplacian_surface_editing}.
%The set of neighboring vertices is denoted as $N(w,h)$.
% \begin{equation}
%     L_m = \sum_{w,h} || \sum_{\overline{w},\overline{h} \in \mathbb{N}(w,h)}(V_{\overline{w}, \overline{h}} - V_{w, h})||_1
% \end{equation}
% \vspace{-.1in}

\subsubsection{Geometry Aware Multi-View Relighting}
% \subsubsection{Multi-View Relighting Using A Geometry-Aware Network - MVR}
\label{subsubsec:mvr}
\vspace{-.1in}
Encoding the scene geometry priors and the relationship between scene geometry and lighting effects is an established method of providing strong signals to shadow removal and synthesis networks \cite{mildenhall2020nerf, zhang2021nerfactor, philip2019multi}. The image relighting networks in SIMBAR follow MVR \cite{philip2019multi}, in which a set of geometric priors are leveraged as inputs in addition to the source image. A set of input buffers, $I_B$, are generated which consists of normal maps, reflectance maps, and refined shadow maps. The normal map encodes the surface normals at each pixel. The reflectance map is a dot product between the surface normals and sun directions. To obtain refined shadow maps, a set of coarse RGB shadow maps are used as inputs to two shadow refinement networks - one each for the source and target lighting condition. These coarse RGB shadow map are created from rays cast onto a 3D mesh of the scene to generate shadow locations. For each ray that intersects the mesh and casts a shadow, let $m_i$ represent the point of intersection. The coordinates of $m_i$ can be re-projected to find the corresponding 2D image pixel and its RGB value. The latter is encoded in the shadow maps to create RGB shadow maps. Encoding the RGB value that corresponds to the object that cast the shadow can help the shadow refinement networks correct the errors made by the 3D mesh reconstruction, in order to produce final refined shadow maps for the relighting network.
% Nikita: Focal length estimation is used, and the camera pose is centered at the origin relative to the mesh coordinates due to the depth-based sheet warping.

To finish the relighting process, a third network is used in combination with the shadow refinement networks. All of them are pre-trained on synthetically rendered data. Given the input image and RGB shadow maps for the source and target lighting conditions, the source and target shadow refinement networks attempt to refine the shadow maps to correct for errors in the mesh construction. This is followed by the final relighting network that takes in both the scene priors and refined shadow maps to produce the relit output.

% \vspace{-.05in}
\subsection{Method Description: SIMBAR}
\label{subsec:method}
% \vspace{-.1in}
% which is not generally available owing to its setup requirement
Most prior scene relighting methods \cite{srinivasan2021nerv,zhang2021nerfactor,philip2019multi} require multiple images with different viewpoints. In contrast, SIMBAR leverages monocular depth estimation to obtain geometry approximation. SIMBAR is modular with two distinct components, geometry estimation and image relighting. The full pipeline is shown in Fig.~\ref{fig:simbar_pipeline}. The geometry estimation module (a) represents the scene as a 3D mesh, which allows for a variety of informative priors to be generated for the image relighting module (b). 
% In terms of the overall pipeline structure, SIMBAR is closest to MVR. 
This allows for a novel system design of a single image-based scene relighting that leverages explicit geometric scene representations. 
% \vspace{-.09in}

\subsubsection{Geometry Estimation Component}
\label{subsubsec:geometry_estimation}
% \vspace{-.05in}
The geometry estimation module in SIMBAR generates a 3D scene mesh $M$, from a single input image $I$, as shown in Fig.~\ref{fig:simbar_pipeline}. This is in direct contrast to MVR, which relies on SFM+MVS \cite{schoenberger2016mvs,schoenberger2016sfm} for multi-view scene reconstruction. The steps taken to generate the mesh $M$ from a single image $I$ are inspired by WorldSheet (refer Section~\ref{subsubsec:worldsheet}), but with
additional modifications for improved mesh reconstruction.
% The steps taken to generate the mesh $M$ from a single image $I$ (refer Section~\ref{subsubsec:worldsheet}). 

In SIMBAR, an external pre-trained monocular depth estimation network is used to provide depth information for generating the scene mesh. This is because higher-quality meshes are given for outdoor driving scenes when leveraging the Worldsheet variant that uses an external depth prediction rather than the full end-to-end pipeline that predicts depth and grid offsets. This observation makes sense as with WorldSheet trained models, there is no direct loss on the mesh $M$ in the end-to-end training regime. The supervision is instead obtained only via rendering losses on the final relit image. Thus, the predicted grid offset may not be as geometrically accurate as the one obtained using an external depth network. In addition, we have adapted new monodepth backbones for improved scene geometry estimation for relighting purpose. 

\begin{figure}[ht]
\centering
\includegraphics[width=0.45\textwidth]{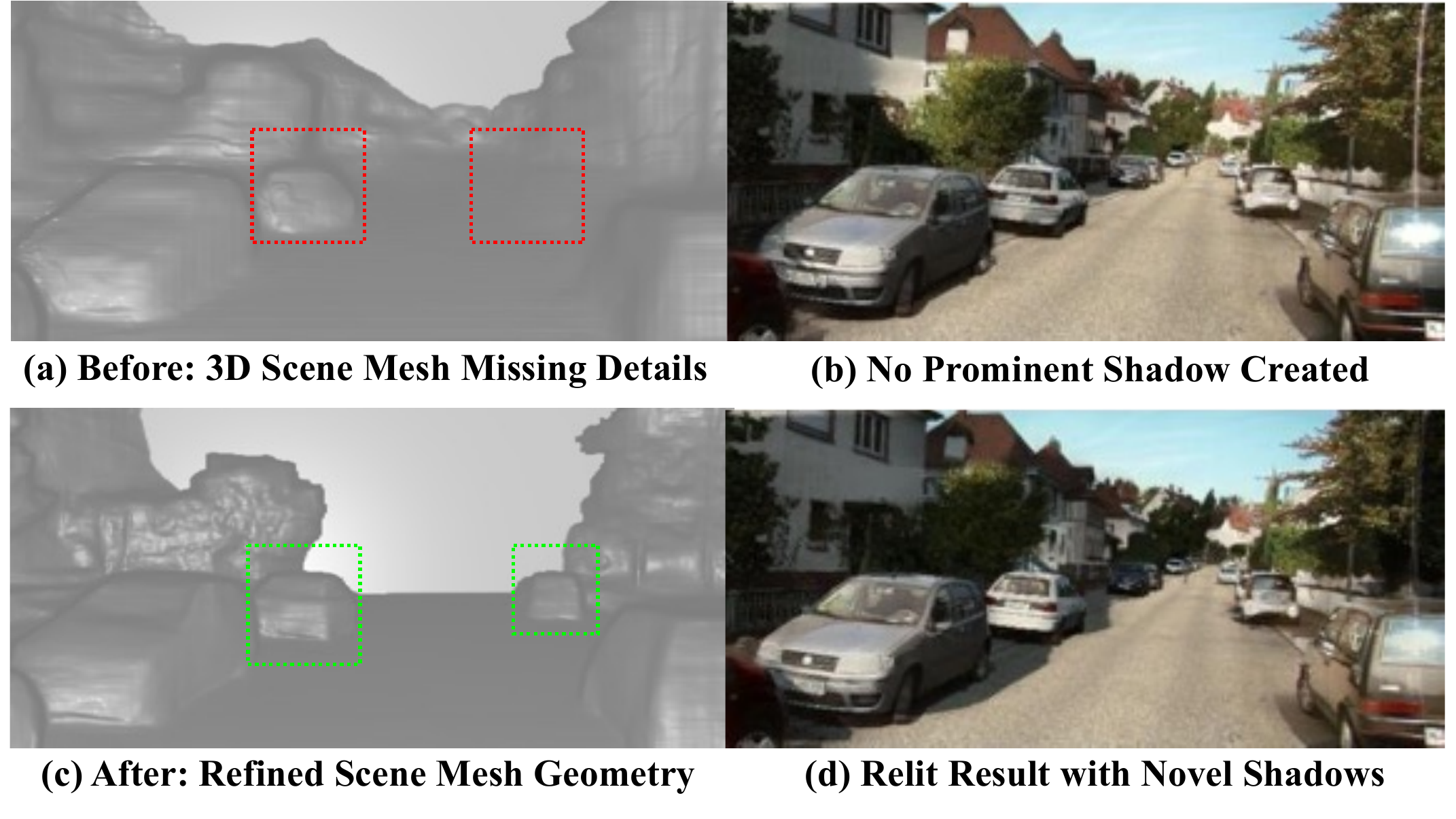}
\vspace{-0.1in}
\caption{(a) With MiDaS v2.1, the 3D scene mesh misses details, resulting in (b) no prominent shadow created. (c) Our improvements with DPT Hybrid which leverages dense vision transformers captures far-away car objects, (d) creating realistic shadows.
\label{fig:exp_midasv2_v3}}
\vspace{-.1in}
\end{figure}

\vspace{-.1in}
%  to investigate how depth predictions affect relighting results. 
\textbf{Improved Monocular Depth Estimation:} While WorldSheet utilizes MiDaS v2.1 as the external depth backbone, we have experimented with Dense Prediction Transformer (DPT) monodepth models \cite{ranftl2021vision}). Fig.~\ref{fig:exp_midasv2_v3} shows that the generated mesh $M$ misses faraway car objects with the MiDaS v2.1 depth prediction, thus missing out on encoding structural details that can potentially cast shadows. This issue is particularly visible in the case of the KITTI scene on the top row, where the faraway car objects are not well relit. To address this limitation, we find that using the improved, dense vision transformers in DPT Hybrid-Kitti (finetuned on KITTI), helps produce more detailed meshes.

\begin{figure}[ht]
\centering
\includegraphics[width=0.45\textwidth]{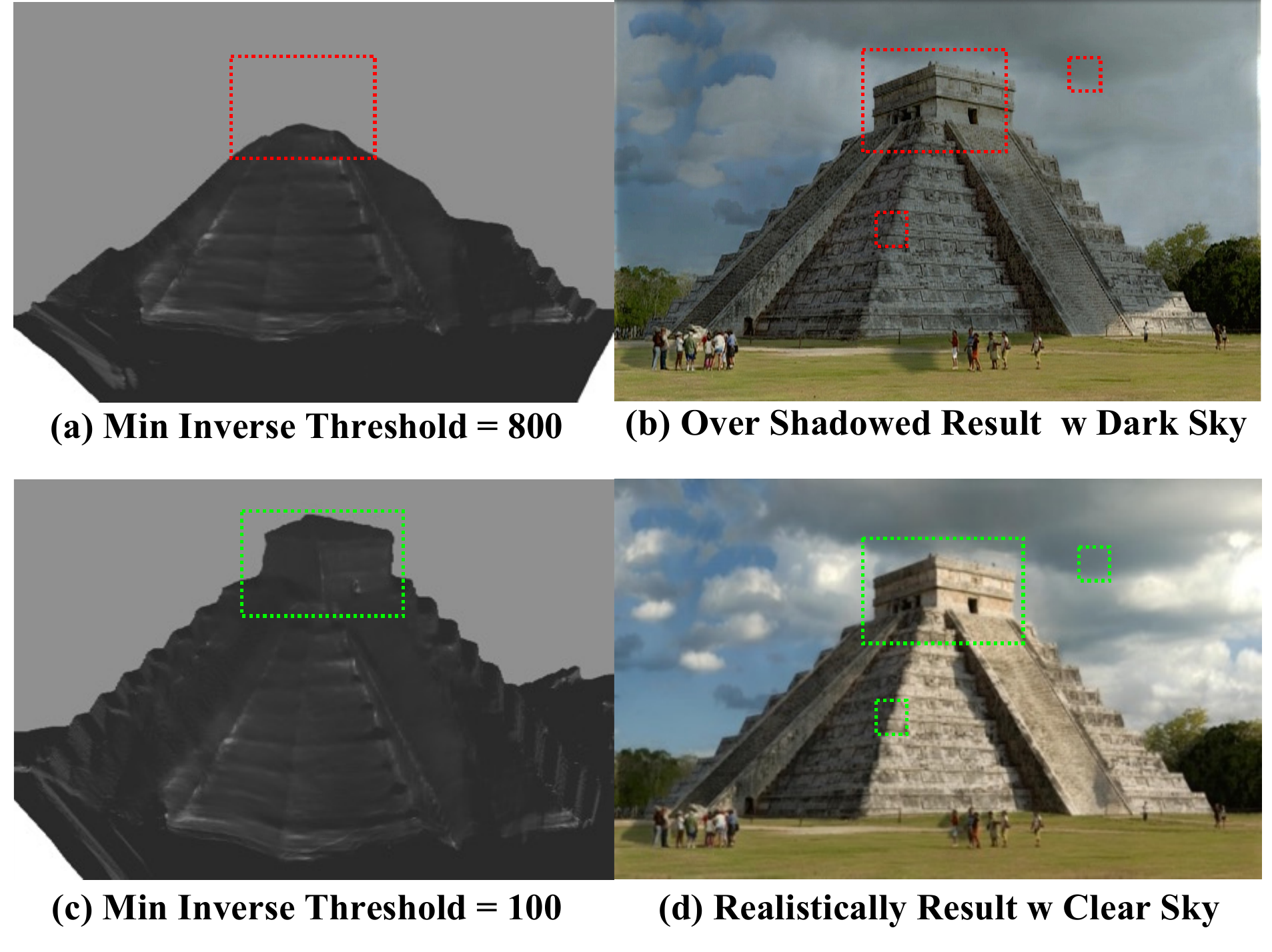}
\vspace{-.2in}
\caption{ (a) For min inverse depth of 800, the scene mesh forms a flat vertical surface at the corresponding threshold distance. This phenomenon is observed as a flat, light gray wall mistakenly cuts off the top part of pyramid geometry. (b) This wall artifact casts a large dark shadow in the relit image labeled ``over shadowed". (c) Using min inverse depth of 100 effective pushes the wall boundary further away, which gives a greater level of detail to the scene mesh, resulting in (d) a more realistic clear shadowed result.
\label{fig:inv_depth_relit}}
\vspace{-.15in}
\end{figure}

\textbf{Foreground/Background Scene Separation:} As shown in Fig.~\ref{fig:simbar_pipeline}, for a given input image $I$, a pre-trained monocular depth estimation network is used to obtain the pixel-wise inverse depth values $D$. These values are then used to inform the deformation of a planar scene mesh. We observe that thresholding the inverse depth at different scales allows us to focus on different levels of detail.

Experiments with different levels of inverse depth thresholds are shown in Fig.~\ref{fig:inv_depth_relit}. For the high inverse depth threshold of 800,  a wall surface is generated fairly close to the camera and scene content. This set up could work for scenes with low depth range, but fail at diverse outdoor scenes with various depth boundaries. This results in over-shadowed results where the fake surface casts its own shadow over the scene. We opt for a lower inverse depth threshold, since this corresponds to a distance further away from the camera position. This allows the mesh to extend further back and produces cleaner shadows. Both the sky and surfaces far away in the horizon are better represented in the mesh $M$ with a lower inverse depth threshold. 
% Note that while the generated mesh is not perfectly consistent with scene geometry, it allows for a general enough understanding of the 3D structure of the scene to generate informative priors to be fed as inputs to the subsequent image relighting module.
% We denote the version of SIMBAR that utilizes low relative inverse depth thresholds as \textbf{S}, as the baseline sigleview approach
%. both scene meshes generated from the same input image are overlayed on top of eachnd kept the same at the foreground for both over-shadowed scene and relit improved scene . Background mesh has been intentionally pushed further away to augment the depth of background objects at far depth range such as sky and road end, to resolve the issue of error casted shadow due to the scene mesh boundary formed at the foreground depth range.
\vspace{-.1in}

\subsubsection{Image Relighting Component}
\label{subsubsec:image_relighting}
\vspace{-.05in}
As shown in Fig.~\ref{fig:simbar_pipeline}, given the scene mesh $M$ from the geometry estimation module, a set of priors or input buffers, as described in Section~\ref{subsubsec:mvr}, are generated. They are fed as inputs to the shadow refinement networks ($r_{src}$, $r_{tgt}$) and the subsequent image relighting network ($r_{out}$). We choose to use MVR's pre-trained networks for $r_{src}$, $r_{tgt}$ and $r_{out}$ since they performed well despite imperfect mesh constructions across different datasets. Furthermore, obtaining a large and diverse set of high-resolution synthetic data for re-training the relighting networks is both time and cost intensive. Therefore, in SIMBAR, we focus on the novel adaption to single-view geometry-aware scene relighting.

\subsection{Improved MVR Medthod as Baseline: MVR-I}
\label{subsec:improved_mvr}
\vspace{-.05in}
The out-of-the-box MVR method fails at single-view collected autonomous driving dataset. To allow for comparisons with a strong baseline, we optimize MVR for road driving scenes with limited views, which we call MVR-I. We use MVR-I as a baseline for all qualitative (Section~\ref{subsec:relit_results}) and quantitative comparisons (Section~\ref{subsec:quant_comparison}).
\vspace{-.1in}
\begin{figure}[ht]
\centering
\includegraphics[width=0.47\textwidth]{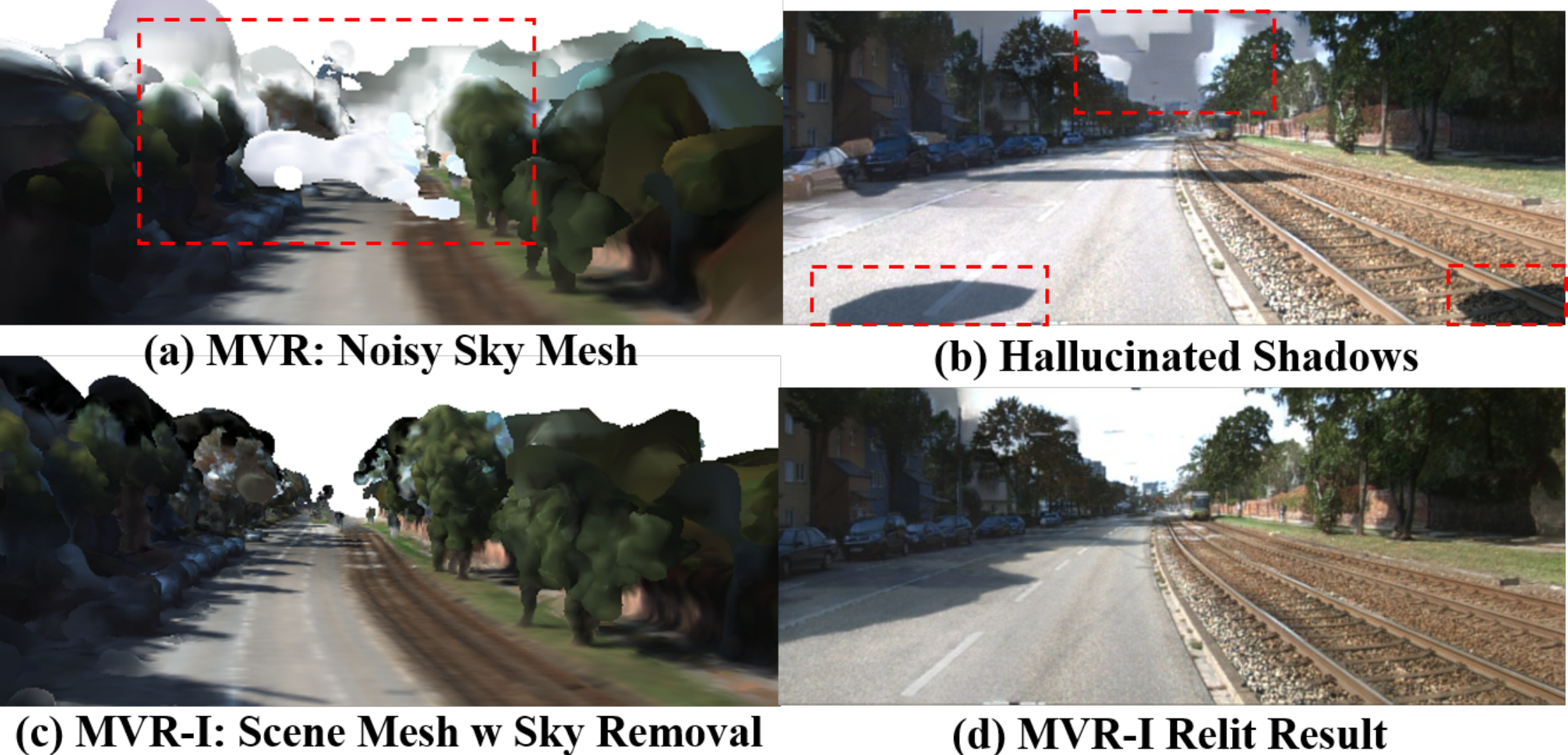}
\vspace{-.1in}
\caption{\small RGB point cloud overlayed on top of the scene mesh generated for a KITTI scene visualization. Using out-of-the-box MVR, hallucinates surfaces in the sky (a) resulting in phantom shadows (b), which we improve with MVR-I (c) leading to a more realistic image relighting result (d). 
\label{fig:exp_improve_mvr}}
\vspace{-.1in}
\end{figure}

\textbf{Removal of Hallucinated Mesh Surfaces:} Firstly, we find that running MVR on KITTI scenes results in hallucinated sky surfaces in the generated mesh, thus casting corresponding phantom shadows on the ground. This is because SFM+MVS reconstruction triangulates selected 3D feature points in the input images with low re-projection error across images. In Fig.~\ref{fig:exp_improve_mvr}, note that the triangulated points leading to surface reconstruction in the sky in (a). These hallucinated surfaces cast prominent shadows in the sky and also on the foreground corner in the relit image in (b). While minor inaccuracies in the mesh can be addressed by the shadow refinement networks \cite{philip2019multi}, the major inaccuracies shown lead to unrealistic scene relighting effects.  To solve this issue, we implement a simple yet highly effective fix. We exclude confounding factors that appear in the sky in (c), such as clouds, as well as the sky itself, through segmentation using Detectron2 \cite{wu2019detectron2} on the input multi-view images. This addresses the issue of hallucinated mesh surfaces in the sky and corresponding phantom shadows (d).
% The top left image shows the RGB point cloud projected from the input image, overlaid on top of the generated scene mesh. % are of the same color as the sky and clouds in the input RGB image, i.e.\ blue and gray. 

\begin{figure}[ht]
\centering
\includegraphics[width=0.44\textwidth]{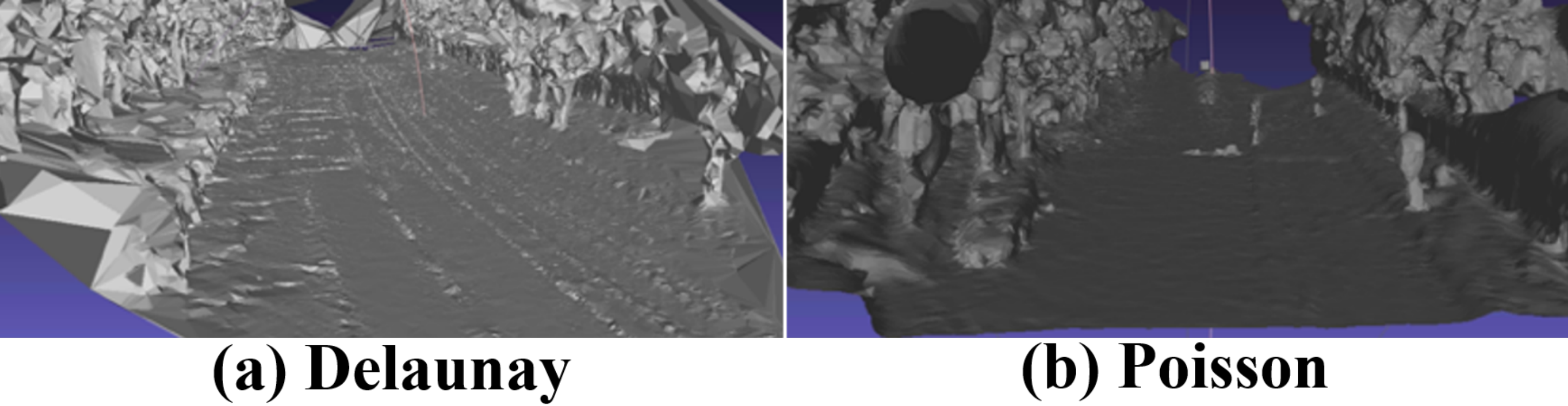}
\vspace{-.1in}
\caption{\small (a) Delaunay surface reconstruction is sensitive to noise, causing triangular artifacts. (b) Poisson reconstructed mesh has much smoother surfaces.
\label{fig:exp_poission}}
\end{figure}
\vspace{-.1in}

\textbf{Improved Surface Reconstruction:} The second improvement is replacing the Delaunay surface reconstruction algorithm \cite{cazals2006delaunay} for mesh generation with the Poisson surface reconstruction algorithm \cite{kazhdan2006poisson}. Fig.~\ref{fig:exp_poission} (left) shows that the Delaunay algorithm results in a noisy mesh, especially for the ground surface. Poisson surface reconstruction (right) for the same scene results in fewer angled edges and overall smoother road and tree surfaces. 
% \todo{Replace this figure with RGB pixels overlaying on top.}

A natural outcome of both these fixes is more realistic relighting results, as shown in Fig.~\ref{fig:exp_improve_mvr} (d).

% \begin{figure}[ht]
% \centering
% \includegraphics[width=0.45\textwidth]{latex/figures/section4_experiments/exp_after_sky_removal.png}
% \caption{\small Improved MVR method. Semantic segmentation used for scene object classfication (top left),  following with confounding object classes removal (bottom left), as a result, scene mesh of road driving scene no longer contain noisy sky points (bottom right), thus corresponding relit results will not be affected by hallucinated shadows.
% \label{fig:exp_after_sky_removal}}
% \end{figure}

% Different Depth Models Benchmarked 
% MiDAS v2.1
% MiDAS v3 Hybrid
% MiDAS v3 Hybrid-Kitti
% MiDAS v3 Large, 21% improvements

% \begin{algorithm}[ht]
% \small
% \caption{\small SIMBAR} \label{alg:relighting}
% \hspace*{\algorithmicindent} \textbf{Input:} Single RGB Image $I$, Source Lighting Direction $s_{src}$, Target Lighting Direction $s_{tgt}$ \\
% \hspace*{\algorithmicindent} \textbf{Output:} Single Relit RGB Image $R$
% \begin{algorithmic}[1]
% \State $z_{w,h} \gets \texttt{MiDaS}(I) ~\forall~ w,h \in I$ 
% \State $\min(z_{w,h}, \beta) ~\forall~ w,h $
% \State $V_{i, j} \gets \texttt{Eq.1}(z_{w,h}, x_{w,h} \theta_f)$ \Comment{ $V \in \mathbb{R}^{129} \times \mathbb{R}^{129}$}
% \State $L_m \gets \texttt{Eq.2}(V)$ \Comment{Mesh face generation}
% \State $B, M \gets g(L_m)$ \Comment{$g: \langle V,L_m \rangle \rightarrow \mathbb{R}^{5} \times \mathbb{R}^{I}$ }
% \State $r_{out} \gets r_{in}(I, M_{src}), r_{tgt}(I, M_{tgt}), B$
% \end{algorithmic}
% \end{algorithm}

\subsection{Scene Relighting Results}
\label{subsec:relit_results}
\begin{figure}[ht]
\centering
\includegraphics[width=0.45\textwidth]{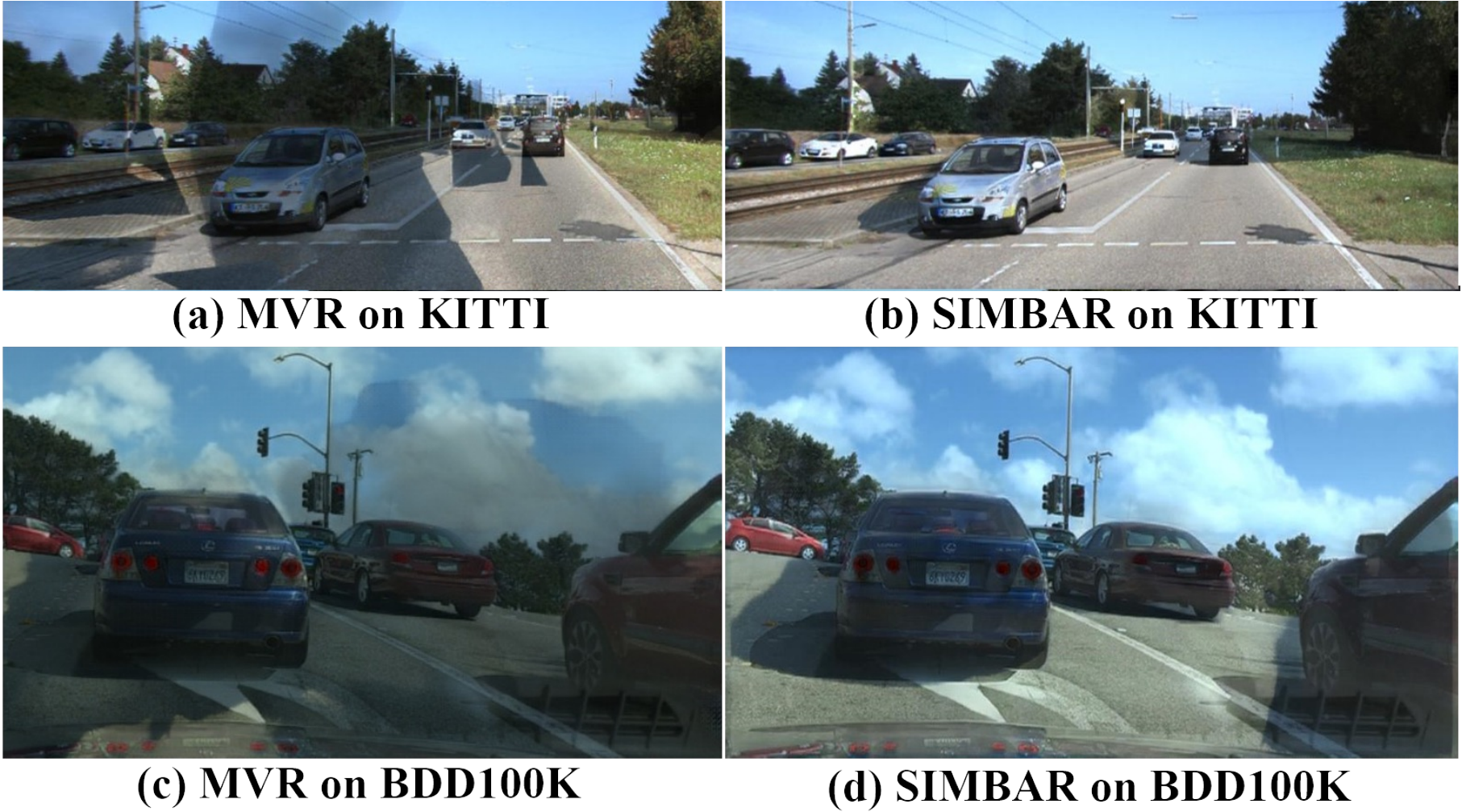}
\vspace{-.1in}
\caption{\small Relighting results from MVR-I (a)(c) and SIMBAR (b)(d) on KITTI and BDD100K respectively.
\label{fig:exp_mr_sr_1}}
\end{figure}

\vspace{-.05in}
Both MVR and MVR-I require multiple viewpoints of a scene to generate an approximate 3D mesh using SFM+MVS. Such an approach fails in video sequences captured by a stationary ego vehicle because of the lack of multiple view-points within the captured sequence. This is a known limitation of SFM+MVS, which leads to many hallucinated shadows rendered in a KITTI frame relit using MVR-I. This can be observed in the top row in Fig.~\ref{fig:exp_mr_sr_1}. 
% An additional scenario in which MVR-I does not work well is when video sequences are captured with minimal feature overlap between frames.
\vspace{-.1in}
% \vspace{-.1in}
In contrast, SIMBAR provides significantly more realistic and geometrically consistent relighting results, as shown in Fig.~\ref{fig:exp_mr_sr_1} (b) and (d). While MVR-I fails to realistically relight images of road driving scenes from both KITTI (top) and BDD100K (bottom), SIMBAR's relighting results are consistently more realistic in terms of target shadow orientations and sky colors. However, there are some strong cast shadow residual that cannot be removed cleanly.

\subsection{Limitations}
\vspace{-0.05in}
% %original
% With our proposed improvements over WorldSheet in the geometry estimation module (as described in Section~\ref{subsubsec:geometry_estimation}), there is significant improvement in the generated mesh leading to better representation of surface details of foreground objects and better inclusion of background objects (refer Fig.~\ref{fig:inv_depth_relit} and Fig.~\ref{fig:exp_midasv2_v3}). However, there is still scope for further improvement to obtain a more explicit representation of small objects present at very far depth range in the scene.

%new
\textbf{Full Occlusion}: With our proposed improvements in the geometry estimation module (refer Section~\ref{subsubsec:geometry_estimation}), there are significant improvements in the generated mesh leading to more surface details of foreground objects and better inclusion of background objects. However, the natural drawback of a monocular depth approach is the exclusion of fully-occluded objects. While partially-occluded objects mesh errors can be corrected by shadow refinement networks, fully occluded objects currently present issues with shadow removal. The mesh is unable to represent the object without an additional view containing the object, yet in the real input image, the object can still contribute shadows. We find this to occasionally result in shadow residue from shadow removal due to the lack of context on the object when utilizing single-view sources.

\textbf{Scene Mesh Manipulation}: Using low inverse threshold generates sky objects as wall surface further away in the horizon (see Fig.~\ref{fig:inv_depth_relit}), and ideally we hope to remove the flat wall surface via scene mesh manipulation for more robust scene mesh separation. To achieve a better geometric understanding of individual objects in the scene and more granular control for scene relighting and shadow manipulation, another optimization could be leveraging 3D mesh predictions using neural networks such as Mesh R-CNN\cite{mesh_rcnn}. We currently use the 3D mesh as a geometric representation, and do not model the specific surface properties. Further modeling could allow for realistic lighting effects that account for specular reflection.

% \todo talks about scene lighting as a whole vs individual objects relighting

% Realistic scene relighting cannot be done without a proper understanding of the environment from the 3D geometric representation. Our method SIMBAR only takes a single RGB image, and does not depend on offline software to render the 3D object models in the scene. Specifically, generating data with SIMBAR can be automated for fast production with no manual tuning. Furthermore, SIMBAR leverages on learning based monocular depth estimation for geometric scene understanding, following with mesh warping with WorldSheet, which can widely apply to any unstructured image datasets.

\section{Data Augmentation Using Scene Relighting For Object Detection \& Tracking}
% \vspace{-0.1in}
\label{sec:augmentation}
A serious limitation of all prior works on scene relighting is the lack of quantitative metrics to validate the effectiveness of scene relighting as a useful data augmentation methodology for vision tasks. In the absence of such a metric, the efficacy of real-world applicability of any scene relighting pipeline cannot be assessed. Therefore, to validate the effectiveness of scene relighting as a data augmentation strategy for vision tasks, we present real-world application results by integrating a state-of-the-art simultaneous object detection and tracking model, CenterTrack, with SIMBAR-augmented datasets. Our goal is to evaluate the enhanced generalization capability of vision models trained on data augmented using SIMBAR.

\subsection{Experiment Setup}
\label{subsec:experiment_setup}
\begin{figure}[ht]
\centering
\includegraphics[width=0.47\textwidth]{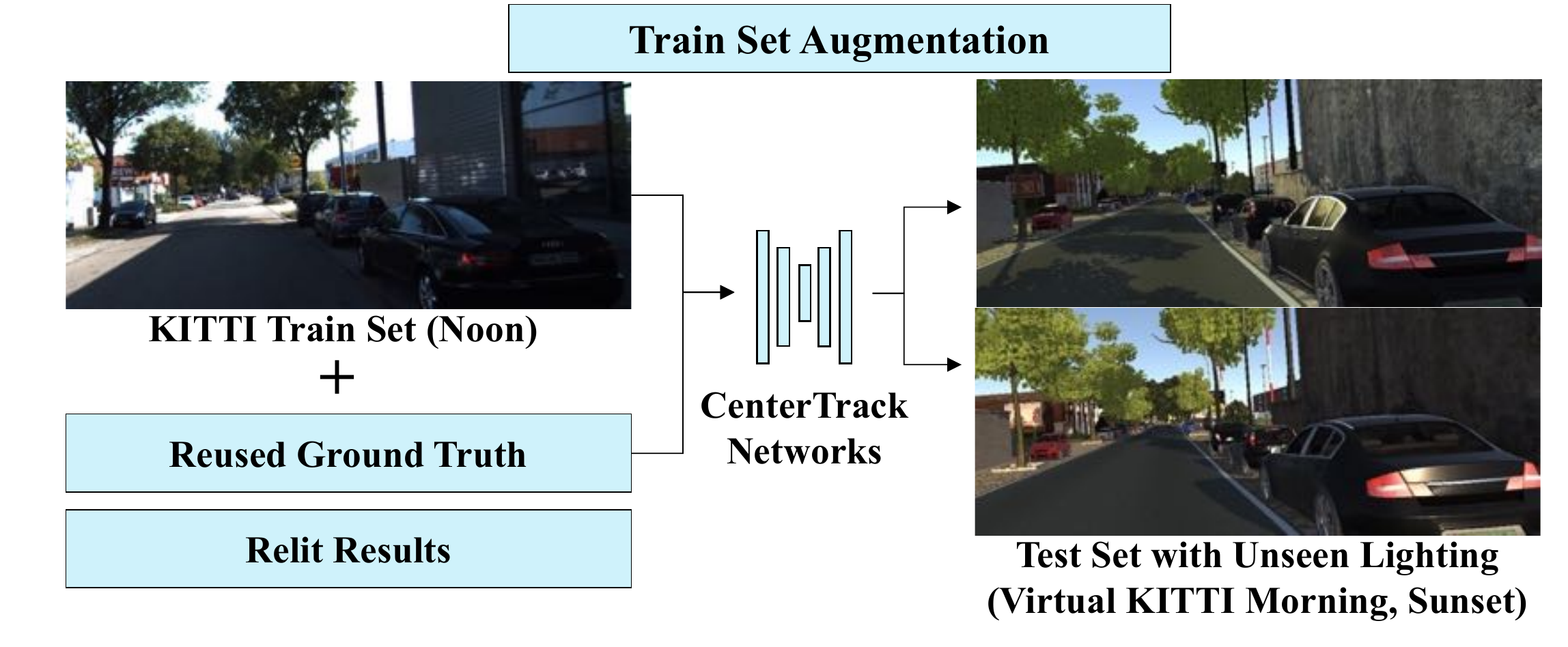}
\vspace{-.15in}
\caption{\small KITTI images taken at noon augmented with MVR-I/SIMBAR relit results used for training CenterTrack models, and vKITTI `morning' and 'sunset' images used for testing. 
\label{fig:exp_unseeen_lighting}}
\vspace{-.1in}
\end{figure}

% \vspace{-0.1in}
\textbf{Train \& Test Datasets:} The KITTI tracking dataset consists of 21 sequences of road scenes, collected during daytime, with minimal variation in lighting conditions. Vision models trained on such a limited dataset cannot generalize well to the wide variety of lighting conditions that might be encountered in the real-world. To approximate this real-world generalization challenge, we train CenterTrack models on KITTI and test on vKITTI (only contains `car' annotations) \cite{gaidon2016virtual} which comprises of `morning' and `sunset' lighting variations. Prior work has also shown that testing on vKITTI is a useful strategy for evaluating data augmentations \cite{tremblay2018training}. A visual of the domain gap between the training and test sets is shown in Fig.~\ref{fig:exp_unseeen_lighting}. Such an experiment setup is important in highlighting that vision models trained on limited datasets are susceptible to failure when encountering a seen scene in unseen lighting conditions.

% \vspace{-.1in}
\textbf{Data Augmentation Using Scene Relighting:} To compare the data augmentation effectiveness of SIMBAR with that of MVR-I (see Section~\ref{subsec:improved_mvr}), we compare the performance of CenterTrack models trained on two types of augmented KITTI datasets. Both use the full training set of ground-truth KITTI sequences, and an augmented versions of sequence numbers: 0001, 0002, 0006. The two augmented datasets differ in how the images are relit, one using MVR-I and the other using SIMBAR, where input parameters such as sun direction and sky zenith are randomly initialized. For our experiments, 4 different relit versions were generated for each frame in the 3 KITTI sequences. However, up to 120 different lighting conditions can be generated for each frame. We perform this augmentation offline. The rest of the training procedure follows the original CenterTrack implementation as-is. For brevity, we will refer to the CenterTrack models trained on the original 21 KITTI sequences without any image relighting-based augmentation as \textbf{(K)}; \textbf{(K+M)} and \textbf{(K+S)} denote the models trained with KITTI augmented with MVR-I relit sequences and SIMBAR relit sequences respectively.

\textbf{Metrics:} To quantify the effectiveness of data augmentation using scene relighting for object detection and tracking, we report the Multiple Object Tracking Accuracy (MOTA), MOT Precision (MOTP), Multiple Object Detection Accuracy (MODA), MOD Precision (MODP), complemented with Precision (P), Recall(R), F1 score, False Positives (FP) and False Negatives (FN). 

\subsection{Evaluation Results}
\label{subsec:quant_comparison}
\vspace{-0.05in}
% \textbf{Quantitative Evaluation}.
A summary of the quantitative results is shown in Table.~\ref{tab:quant}. All models trained from scratch, and the best checkpoint for each training run is chosen based on MOTA on the real KITTI validation set. CenterTrack models trained on KITTI augmented with relit KITTI, from either MVR-I (\textbf{K+M}) or SIMBAR (\textbf{K+S}), consistently outperform the baseline CenterTrack model trained on KITTI (\textbf{K}), on all metrics except for MODP. Specifically, the CenterTrack model trained on KITTI augmented with SIMBAR (\textbf{K+S}) has the highest MOTA of 93.3\% - a 9.0\% relative improvement over the baseline MOTA of 85.6\% from \textbf{K}. Similarly, the highest MODA of 94.1\% is also achieved by \textbf{K+S} - again an impressive 8.9\% relative improvement over the baseline MODA of 86.4\% from \textbf{K}. In addition, \textbf{K+S} has the least amount of false positives and false negatives.

\begin{table}[ht]
\centering
\begin{tabular}{|c|c|c|c|c|}
\toprule
& \textbf{K} & \textbf{K+M}& \textbf{K+S} \\ \hline
\midrule
\textbf{MOTA} $\uparrow$ & 85.6\% & 92.0\% & \textcolor{blue}{\textbf{93.3\%}} \\ \hline
\textbf{MOTP} $\uparrow$ & 83.1\% & 83.5\% & \textcolor{blue}{\textbf{83.5\%}} \\ \hline
\textbf{MODA}  $\uparrow$ &  86.4\% & 92.7\% & \textcolor{blue}{\textbf{94.1\%}}  \\ \hline
\textbf{MODP}  $\uparrow$  & 87.6\% &  87.6\% & 87.4\%  \\ \hline
\textbf{Recall} $\uparrow$ &  94.0\% & 96.5\% & \textcolor{blue}{\textbf{96.9\%}}  \\ \hline
\textbf{Precision} $\uparrow$ & 94.4\% & 97.4\% & \textcolor{blue}{\textbf{98.1\%}}  \\ \hline
\textbf{F1}    $\uparrow$  &  94.2\% & 96.9\% & \textcolor{blue}{\textbf{97.5\%}}  \\ \hline
\textbf{False Positives} $\downarrow$   & 283 &    133 &     \textcolor{blue}{\textbf{95}}\\ \hline
\textbf{False Negatives} $\downarrow$  &  302 &    179 &    \textcolor{blue}{\textbf{157}} \\ \hline
\bottomrule
\end{tabular}
\vspace{-.1in}
\caption{\small Compared to baseline CenterTrack, models trained with data augmented using both MVR-I and SIMBAR provide consistently better performance.
\vspace{-.1in}
\label{tab:quant}}
\end{table}
% \vspace{-.1in}

\begin{figure}[ht]
\centering
\includegraphics[width=0.35\textwidth]{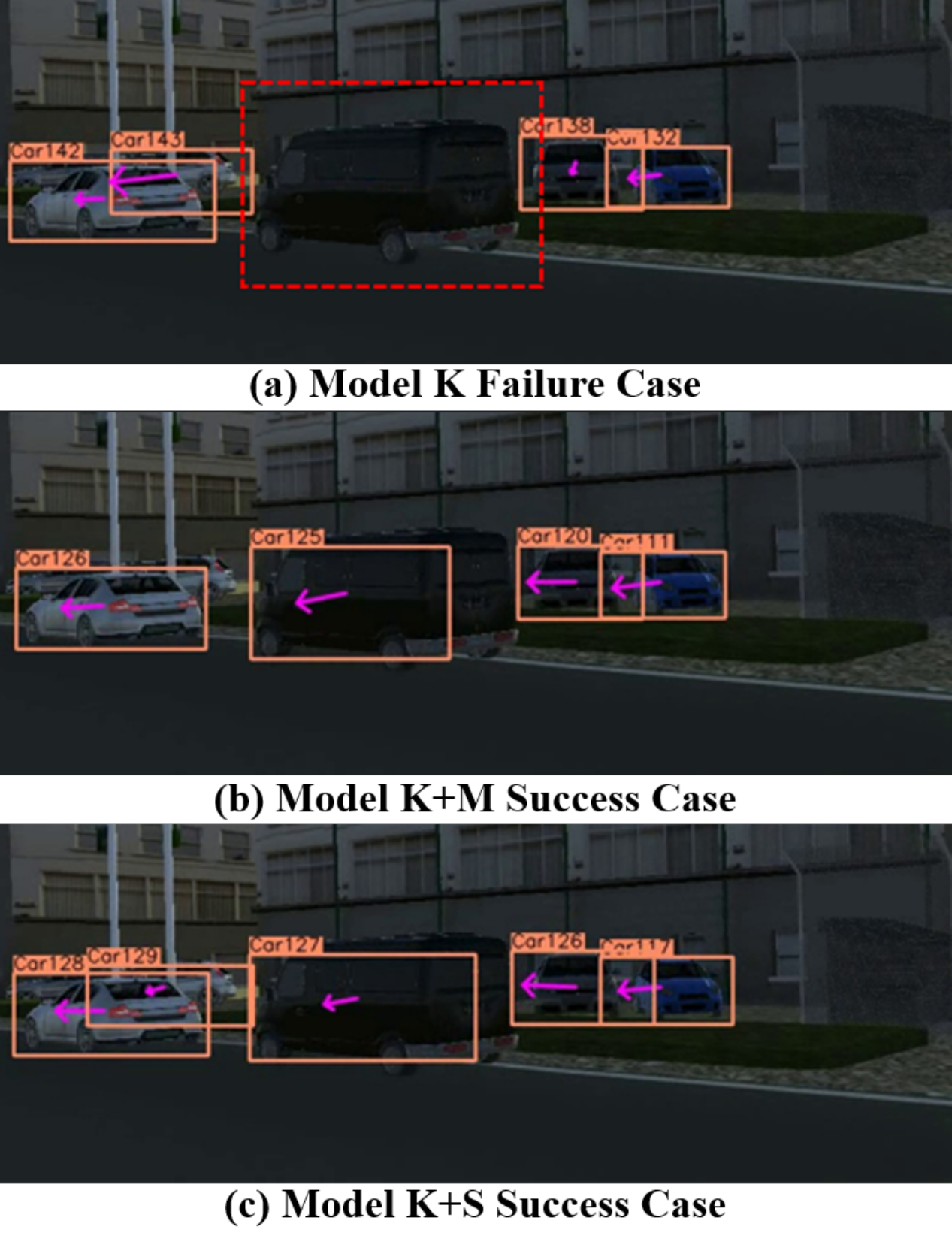}
\vspace{-.2in}
\caption{\small Object detection and tracking results on vKITTI from CenterTrack models \textbf{K} (a), \textbf{K+M} (b), and \textbf{K+S} (c).
\label{fig:exp_quali_1}}
\vspace{-.1in}
\end{figure}

%Should this be a mini section or can we move this to 4.1? If we want to explain the quantitative results here, we should do so without calling this qualitative results. It will cause confusion about what the point of section 4.1 is then. It ends up looking disorganized and incoherent.
% \textbf{Qualitative model performance results}.
% \vspace{-1in}
Fig.~\ref{fig:exp_quali_1} shows a qualitative downstream task performance comparison of detection and tracking results from \textbf{K}, \textbf{K+M} and \textbf{K+S} on vKITTI. The top result shows that model \textbf{K}, trained on original KITTI, fails to detect and track a black van obscured by a dense, dark shadow. Even though \textbf{K} was trained on the exact same scene from KITTI, it fails in this scenario because the training set, limited to images captured at noon, does not contain diverse lighting and shadow variations. Thus, model \textbf{K} performs properly in this seen scene with an unseen lighting condition. In contrast, both the models \textbf{K+M} and \textbf{K+S} perform well on this edge case with challenging lighting conditions.

\begin{figure}[ht]
\centering
\includegraphics[width=0.45\textwidth]{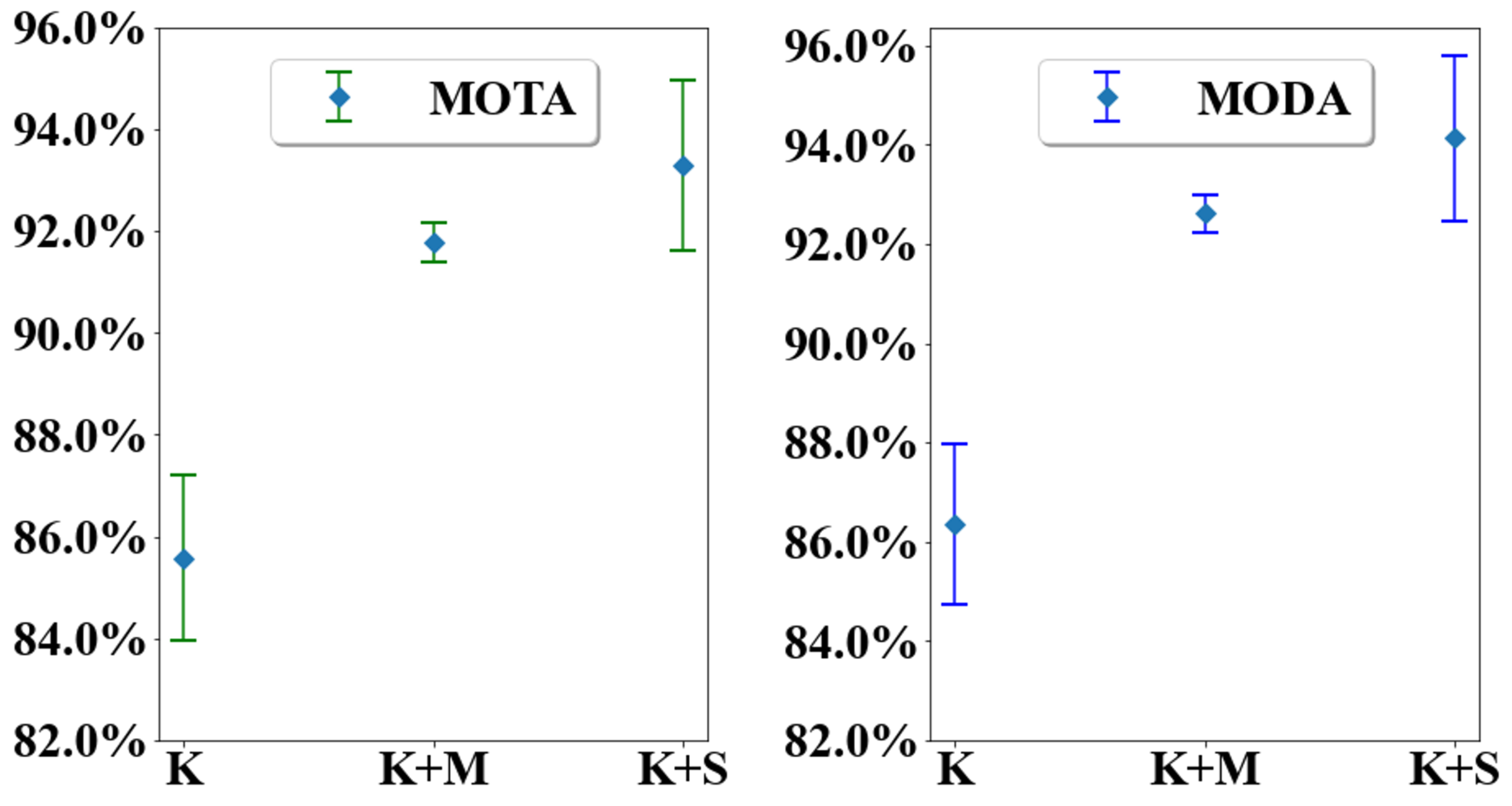}
\vspace{-.1in}
\caption{\small MOTA (left) and MODA (right) variance across 5 different training instances of models \textbf{K}, \textbf{K+M} and \textbf{K+S}.
\label{fig:exp_mota}}
% \vspace{-.3in}
\end{figure}

% \vspace{-.05in}
% \textbf{Variance Measurement}:
%on our internal High Performance Computing cluster$
To investigate the reliability of the obtained improvements, we ran 5 different training instances for each of the three models \textbf{K}, \textbf{K+M} and \textbf{K+S}. Each training instance was run for 100 epochs, which took 9 hours with 8 NVIDIA A100 GPUs. 
Fig.~\ref{fig:exp_mota} shows consistent improvement in MOTA and MODA, averaged across the 5 training runs, with \textbf{K+S} achieving the best overall performance. Note that as compared to \textbf{K}, \textbf{K+M} also performs relatively well on the unseen lighting conditions, with a small variance across different training jobs.

\section{Conclusion}
\label{sec:conclusion}
\vspace{-.1in}
We present a novel single-image based scene relighting pipeline, SIMBAR, for time and cost effective diversification of real-world datasets to include a plethora of lighting conditions. SIMBAR consists of two main modules. The geometry estimation module, inspired by 3D scene geometry estimation from a single image using WorldSheet, exploits various inverse depth thresholds and monocular depth networks to improve the scene mesh. The image relighting module re-purposes the relighting networks from prior art MVR and further relaxes the application-prohibitive requirement of multiple input images with different camera views. An improved version of MVR (MVR-I) is also provided for benchmark purposes. MVR-I leverages segmentation pre-processing to remove confounding classes, and is refined for road driving scenes.

Additionally, a comprehensive quantitative evaluation of CenterTrack models trained on KITTI augmented with relit data is used to demonstrate the effectiveness of scene relighting as a data augmentation strategy for object detection and tracking. Our results show an impressive MOTA of 93.3\% on the vKITTI dataset with CenterTrack trained on KITTI augmented using SIMBAR - a 9.0\% relative improvement over the baseline MOTA of 85.6\% with CenterTrack trained on original KITTI. These results present a strong case for using SIMBAR as an effective data augmentation technique for vision tasks in automated driving.

\textbf{Acknowledgements:} The authors would like to thank Ronghang Hu and Deepak Pathak for sharing WorldSheet source code, and Julien Philip for the inspiring discussions on his MVR relighting method.

%Section~\ref{subsubsec:geometry_estimation} presents our current approach of foreground and background scene mesh separation in details. Although it has improved the scene relighting with realistic rendering results, yet there are still potential to improve the monocular depth estimation for very small scale objects and corresponding relighting results for data augmentation.

%%%%%%%%% REFERENCES
{\small
\bibliographystyle{ieee_fullname}
\bibliography{egbib}
}

\appendix

% --- PDF will be split by an editor (e.g. macOS preview), so need to restart from page 1
% \setcounter{page}{11}

% --- repeat the title (AT: haven't found a more elegant way to do this...)
\twocolumn[
\centering
% \Large
\large
\textbf{Supplementary} \\
\textbf{SIMBAR: Single Image-Based Scene Relighting \\ For Effective Data Augmentation For Automated Driving Vision Tasks } \\    

\vspace{1.5em} 
] 
%< twocolumn
\appendix

\section{Additional Relit Results Comparison}
\vspace{-.2in}
\begin{figure}[ht]
\centering
\includegraphics[width=0.45\textwidth]{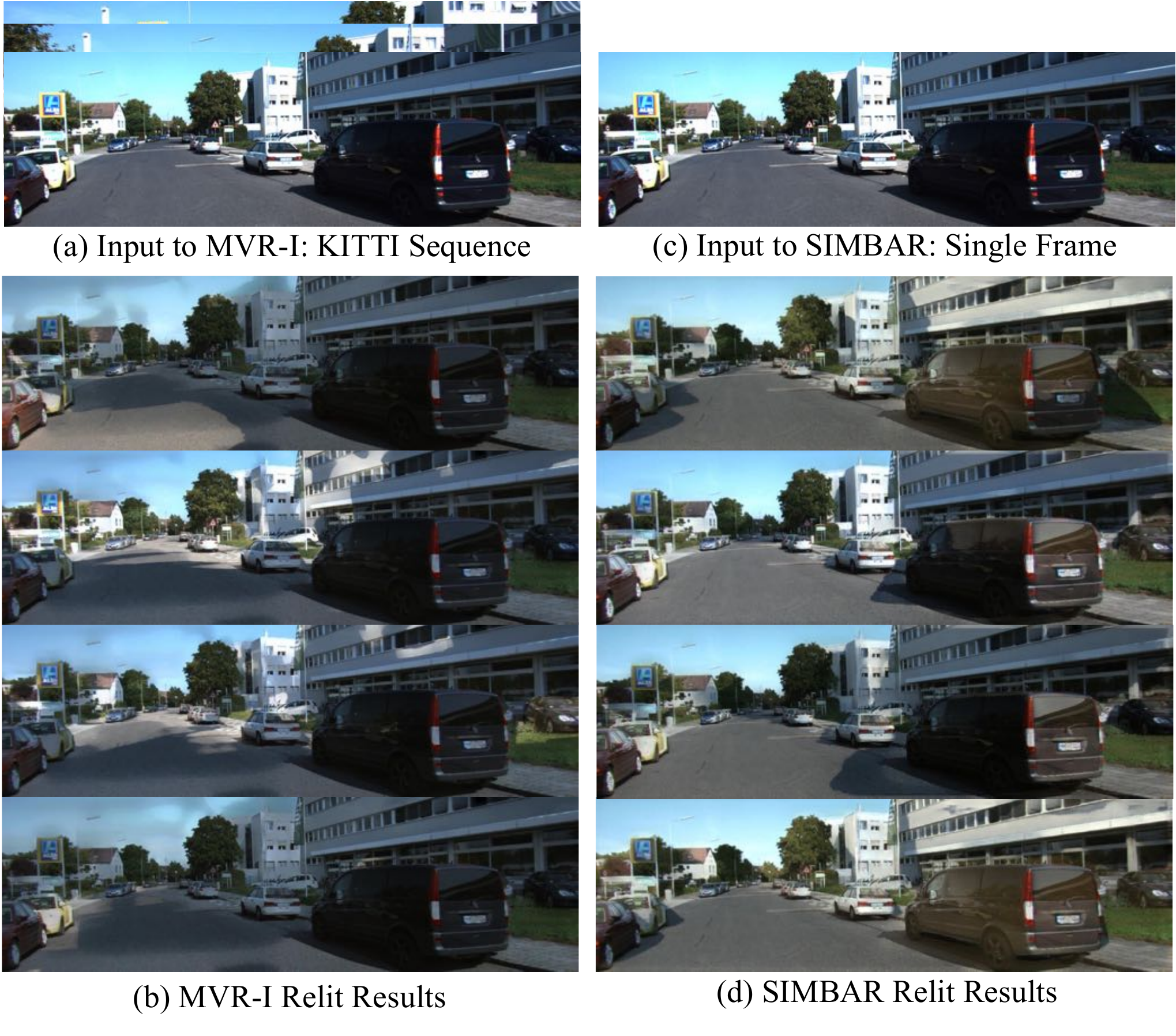}
\vspace{-.1in}
\caption{\small Relit results from MVR-I and SIMBAR on the KITTI traffic intersection sequence 0001.
\label{fig:mvr_simbar_supp_kitti_van}}
\end{figure}
\vspace{-.2in}
\begin{figure}[ht]
\centering
\includegraphics[width=0.45\textwidth]{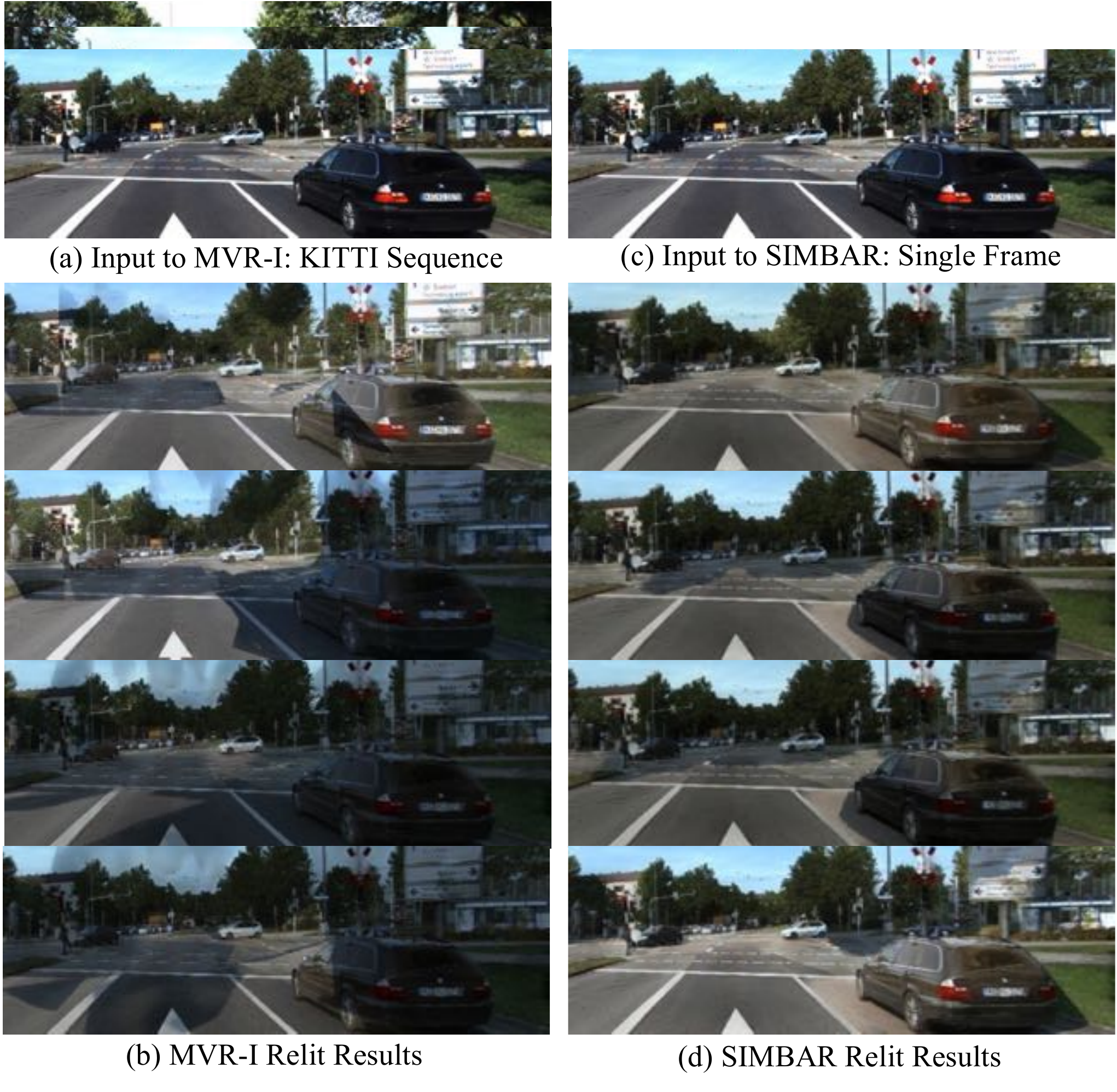}
\vspace{-.1in}
\caption{\small Relit results from MVR-I and SIMBAR on the KITTI road driving sequence 0002.
\label{fig:mvr_simbar_supp_kitti_static}}
\end{figure}
\vspace{-.05in}

\begin{figure}[ht]
\centering
\includegraphics[width=0.45\textwidth]{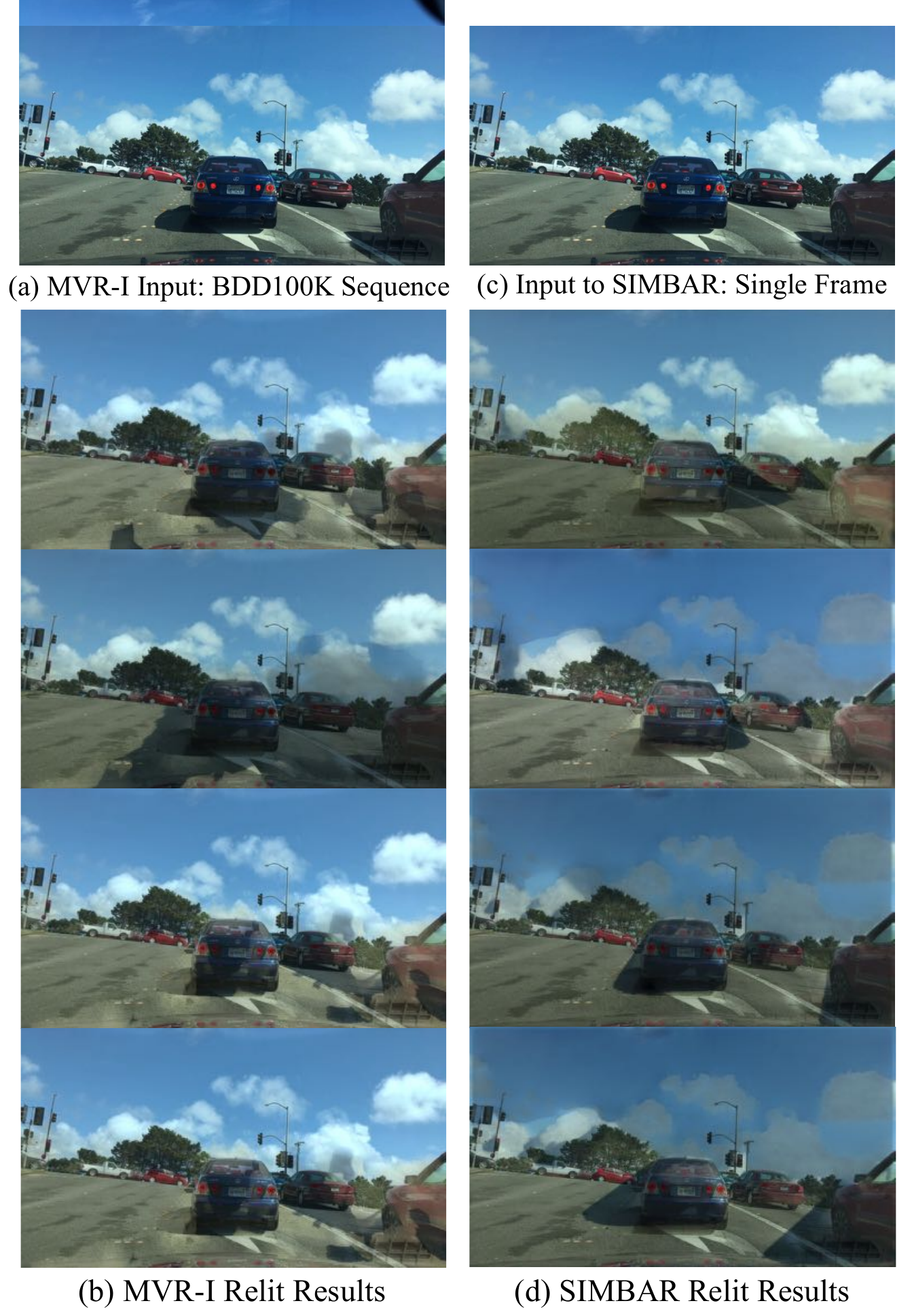}
\vspace{-.1in}
\caption{\small MVR-I and SIMBAR relights the BDD100K traffic intersection.
\label{fig:mvr_simbar_supp_bdd_redlight}}
\end{figure}

\begin{figure}[ht]
\centering
\includegraphics[width=0.45\textwidth]{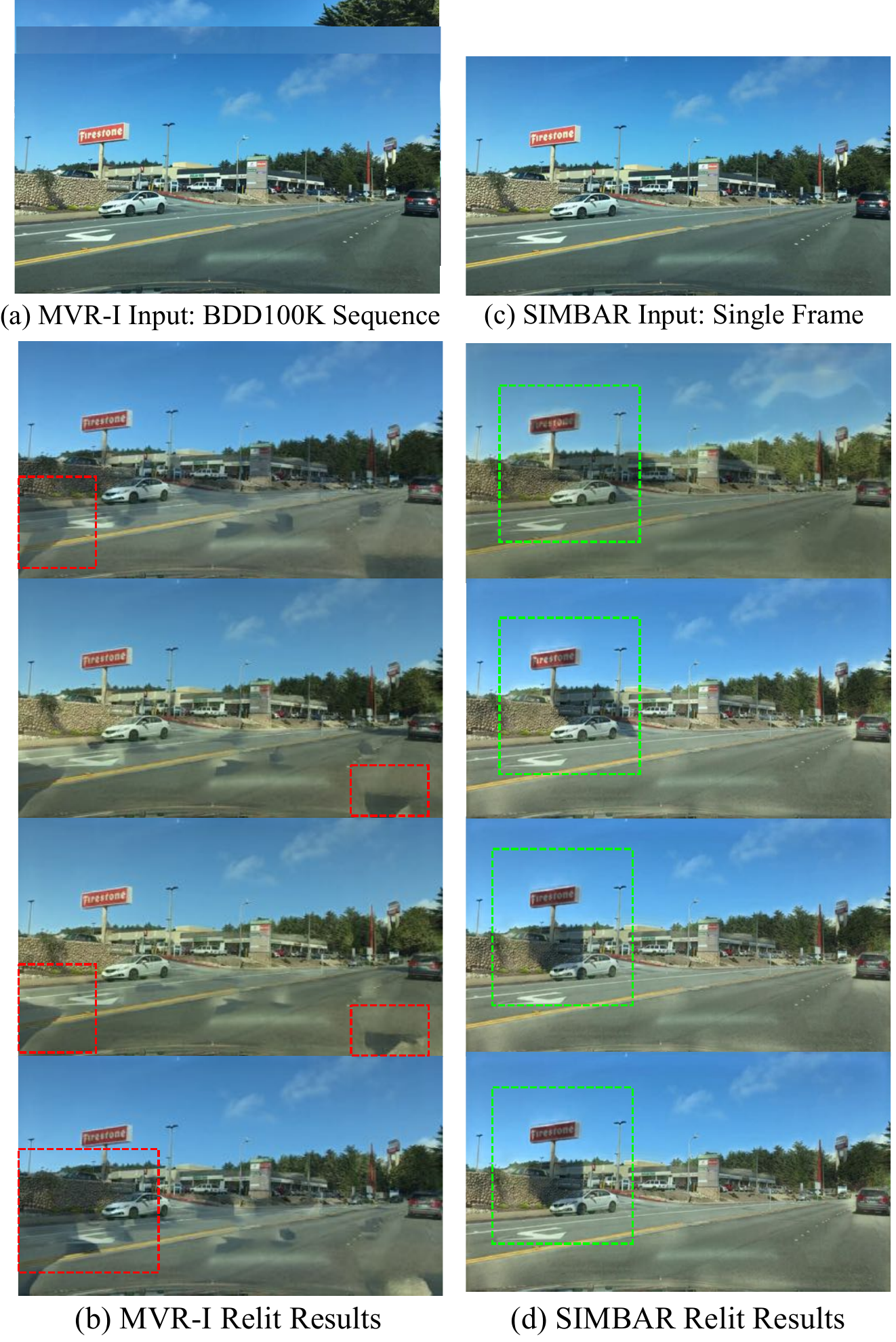}
\vspace{-.1in}
\caption{\small MVR-I and SIMBAR relight the BDD100K road scene with challenging small scale objects.
\label{fig:mvr_simbar_supp_bdd_small}}
\end{figure}
\vspace{-.1in}

\textbf{Qualitative Comparison: } As shown in Fig.~\ref{fig:mvr_simbar_supp_kitti_van}, MVR-I takes the KITTI sequence 0001 as input, and relights the scene based on the mesh reconstruction from multiview input. This frame is captured around a corner, where the MVR-I method created phantom shadows in the scene. Our SIMBAR method generated physically consistent shadows for the car objects.
In Fig.~\ref{fig:mvr_simbar_supp_kitti_static}, the input data is KITTI sequence 0002 captured at a road intersection, where the ego vehicle is static. MVR-I failed to relight this scene with many hallucinated shadows rendered, due to the insufficient multiview information within the captured sequences. In contrast, taking a single frame,  our method SIMBAR provides significantly more realistic and physically consistent relighting results, shown on the right with shadows rendered around the car objects. As highlighted in yellow, SIMBAR relit results also have clear sky. 

In addition, SIMBAR can generalize across different datasets with no need to retrain for every new dataset, as indicated in Fig.~\ref{fig:mvr_simbar_supp_bdd_redlight},  where MVR-I and SIMBAR relight the same BDD100K sequence, with several cars including the ego vehicle waiting for red lights. In contrast, MVR-I again generated phantom shadows in the scene.

\textbf{Corner Case Comparison: } In Fig.~\ref{fig:mvr_simbar_supp_bdd_small}. The input data is BDD100K sequence with very small scale objects in the scene, which is a challenging case for relighting. Neither MVR-I nor SIMBAR could relight two cars in the scene with new shadows. Particularly, MVR-I failed to relight this scene due to the lacking feature matching between frames during the SFM+MVS based mesh reconstruction process, resulting in noisy mesh, which creates hallucinated shadows in the scene (b). SIMBAR relies on depth backbone for geometry estimation, which has captured some background scene objects (billboard and buildings) and created novel shadows for these objects accordingly.

\section{Additional Metrics Evaluation}
\vspace{-.15in}
\begin{table}[ht]
\centering
\footnotesize
\begin{tabular}{|c|c|c|c|c|c|}
\toprule
& \textbf{RMSE} $\downarrow$ & \textbf{PSNR} $\uparrow$ & \textbf{SSIM} $\uparrow$ & \textbf{MS-SSIM}$\uparrow$ & \textbf{UQI}$\uparrow$ \\ \hline
\textbf{MVR-I} & 53.70 & 13.68 & 0.65 & 0.71 & 0.79 \\ \hline
\textbf{SIMBAR} & \textcolor{blue}{\textbf{37.59}} & \textcolor{blue}{\textbf{16.86}} & 
\textcolor{blue}{\textbf{0.69}} &  
\textcolor{blue}{\textbf{0.77}} & 
\textcolor{blue}{\textbf{0.86}} \\ \hline
\end{tabular}
% \vspace{-.1in}
\caption{Scene relighting evaluation on vKITTI Scene 0001.
\label{tab:pixel_metric}}
\end{table}
\vspace{-.1in}
\textbf{Pixel-Based Image Quality Metrics Evaluation} Our paper focuses on the benefits the relighting pipeline presents for downstream tasks with the experiments to demonstrate practical applications of the data augmentation. As shown in Table.~\ref{tab:pixel_metric}, we also include image quality metrics to highlight the quality of SIMBAR's augmentation and show direct pixel-based evaluation metrics. 

The use of SIMBAR demonstrates improvements across the direct image metrics relative to MVR-I. This indicates that SIMBAR is able to not only outperform MVR-I for training downstream models on augmented data, but that the image quality is also improved. 

These above metrics could be reference information for measuring MVR/SIMBAR generated data quality. However, such metrics have two limitations - (i) need for ground truth images for different lighting conditions; and (ii) ineffective measurement of perceptual similarity \cite{zhang2018unreasonable}. Section~\ref{sec:augmentation} introduces a novel evaluation scheme, that compares deep models trained on datasets augmented with their relit versions against the baseline models on a held-out test set - quantifying the data augmentation effectiveness of SIMBAR for a vision task. Such an evaluation is more reliable in demonstrating real-world applicability of any relighting pipeline. 
\vspace{-.1in}
\section{Societal Impact}
\vspace{-.05in}
 With the application towards single images, SIMBAR allows for widespread use. The geometry estimation module can be potentially disruptive to people's privacy and safety due to its ability to perform open-world modeling. SIMBAR alleviates this concern by making use of a 3D scene mesh representation that is devoid of any personally identifiable information (PII), as shown in the pipeline schematic in Fig.~\ref{fig:simbar_pipeline} of an open street scene. Other ways in which SIMBAR can have potential negative social impacts is through the diversification (via relighting) of images scraped off of the internet (as is the case with DIV2K) with PII information in them (such as people's age, relationship, license plates etc. \cite{liu2021machine}). 
 
 We hope to mitigate this concern by incorporating an obfuscation method. Applying an off-the-shelf obfuscation network to SIMBAR's relit results can be challenging. As indicated in Fig.~\ref{fig:exp_quali_1}, given the wide variety of lighting conditions SIMBAR can generate, an obfuscation network trained with limited lighting conditions might not work well on SIMBAR relit images. Thus, we suggest obfuscating sensitive parts of the image as follows. Step 1: perform object detection on the source images to extract the pixel coordinates with PII. Step 2: obfuscate (using Gaussian Blur [2]) sensitive regions from Step 1 in the relit versions of the source images. This is straightforward since SIMBAR alters scene lighting without altering semantic content.
 
\section{SIMBAR Relights Real-World Datasets}
\begin{figure*}[htp!]
\centering
\includegraphics[width=1\textwidth]{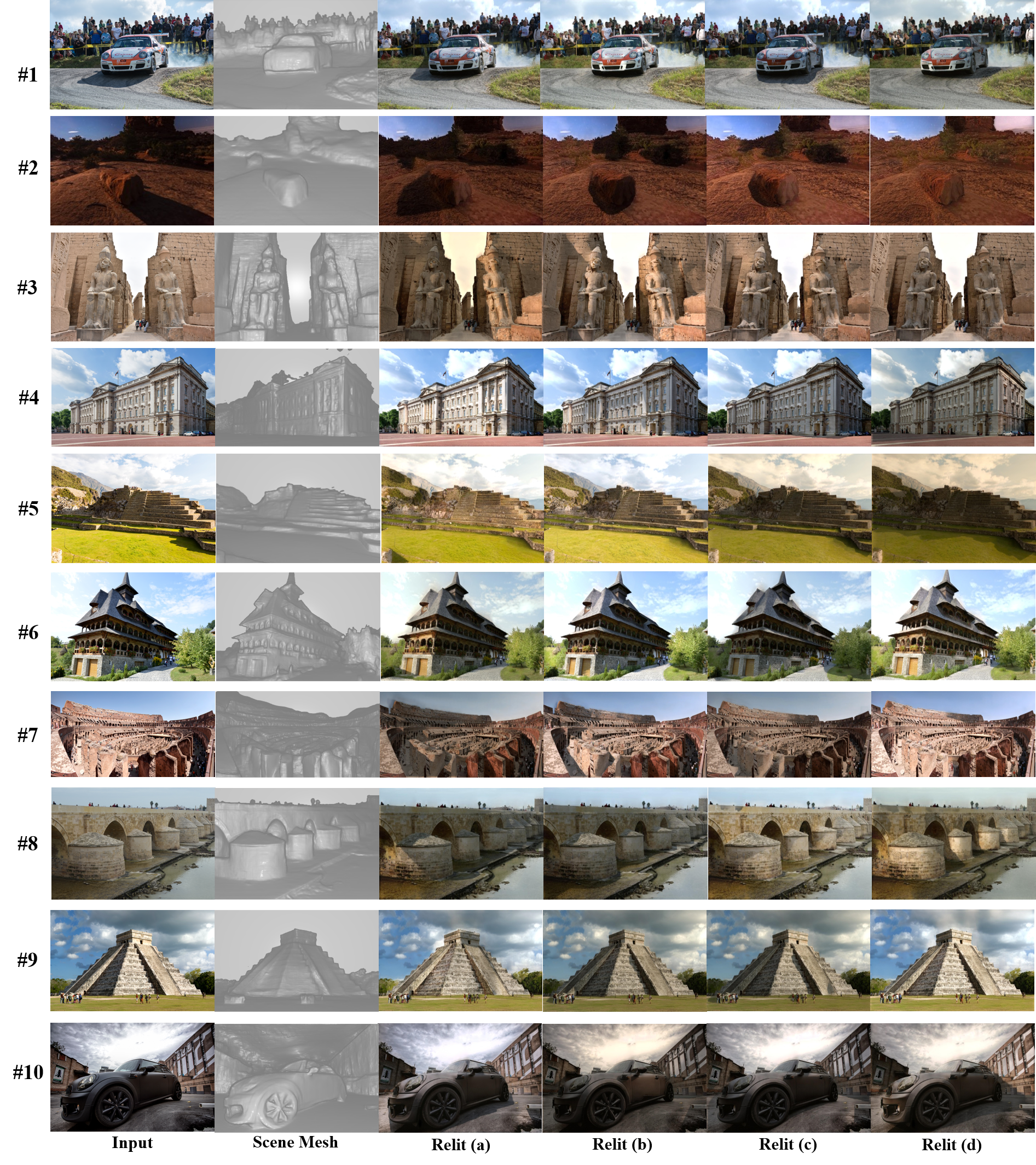}
\caption{\small \textbf{SIMBAR In The Wild:} SIMBAR relights DIV2K Internet Images with wide variety of outdoor scenes. The 1st column shows the input frames, and 2nd column presents the scene mesh generated for geometry estimation. Novel relit results are presented in Relit (a)(b)(c)(d) with 4 different lighting variations. Note that while the generated mesh is not perfectly consistent with scene geometry, it allows for a general enough understanding of the 3D structure of the scene to generate informative priors to be fed as inputs to the subsequent image relighting module. Overall, SIMBAR In The Wild demonstrates our method's generalization capability without retraining for every new dataset. 
\label{fig:simbar_div2k_supp}}
\end{figure*}

\begin{figure*}[ht]
\centering
\includegraphics[width=1\textwidth, height=1\textheight,keepaspectratio]{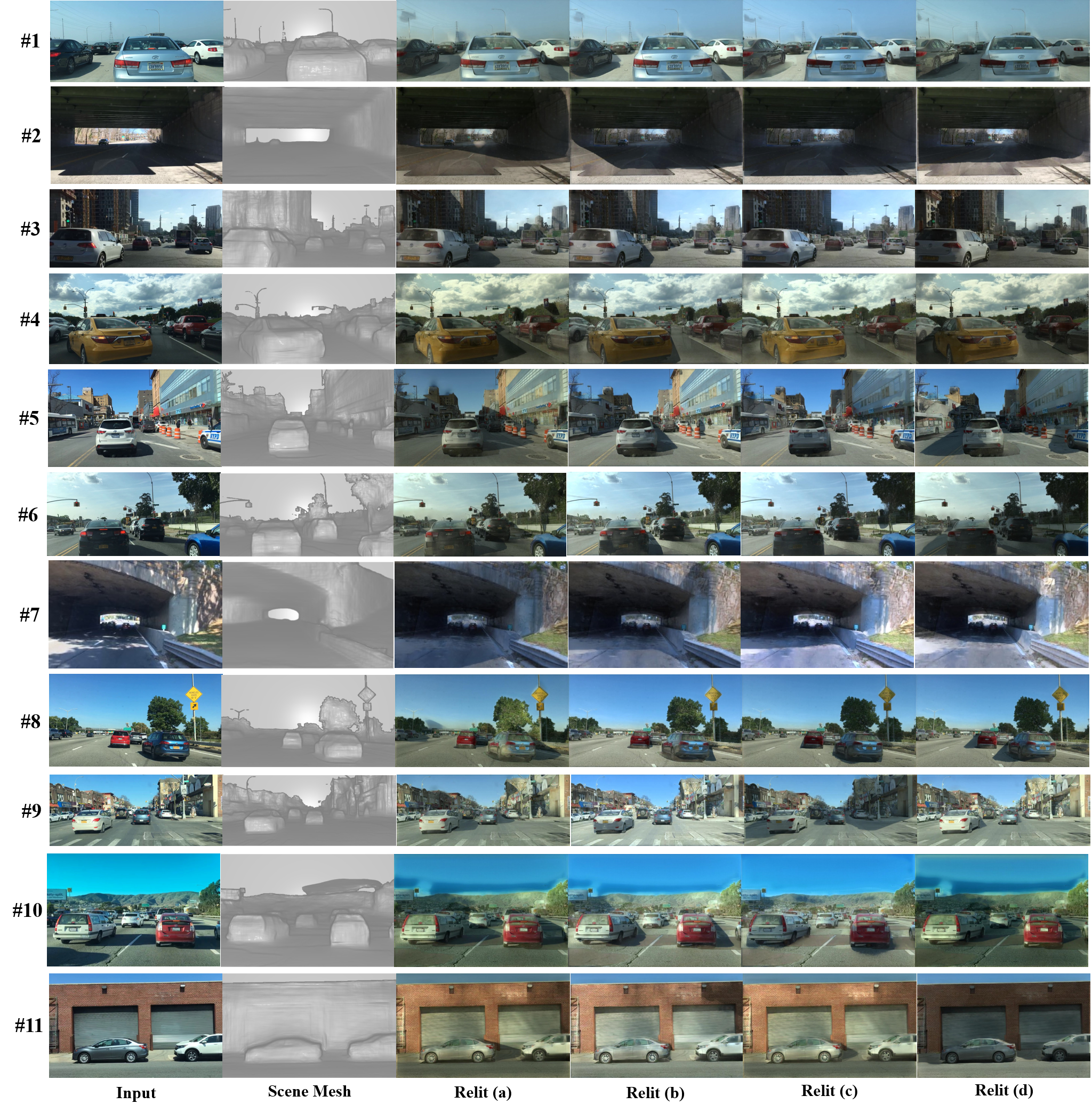}
\caption{\small \textbf{SIMBAR On BDD100K:} SIMBAR relights BDD100K road driving scenes. BDD100K could be a challenging dataset to be relit due to its set up given its large scale, crowd-sourced data collection methodology,and the existence of edge and corner cases. Once again we see robust mesh generation here including in the very complex case of tunnels(row 2 \& 7). The relighting is very effective in these scenes although there is slight warping of the textures. The 3rd row is also interesting because it is an overcast scene so there are no sharp shadows. SIMBAR is able to relight the scene to have sharp shadows but it is not able to raise the overall brightness of the scene. Note that high level scene semantics, such as the clouds, are preserved during relighting, something less controllable methods (e.g. GANs) could struggle to maintain without explicit labels.
% \todo{Review description} 
\label{fig:simbar_bdd100k_supp}}
\end{figure*}

\begin{figure*}[ht]
\centering
\includegraphics[width=1\textwidth, height=1\textheight,keepaspectratio]{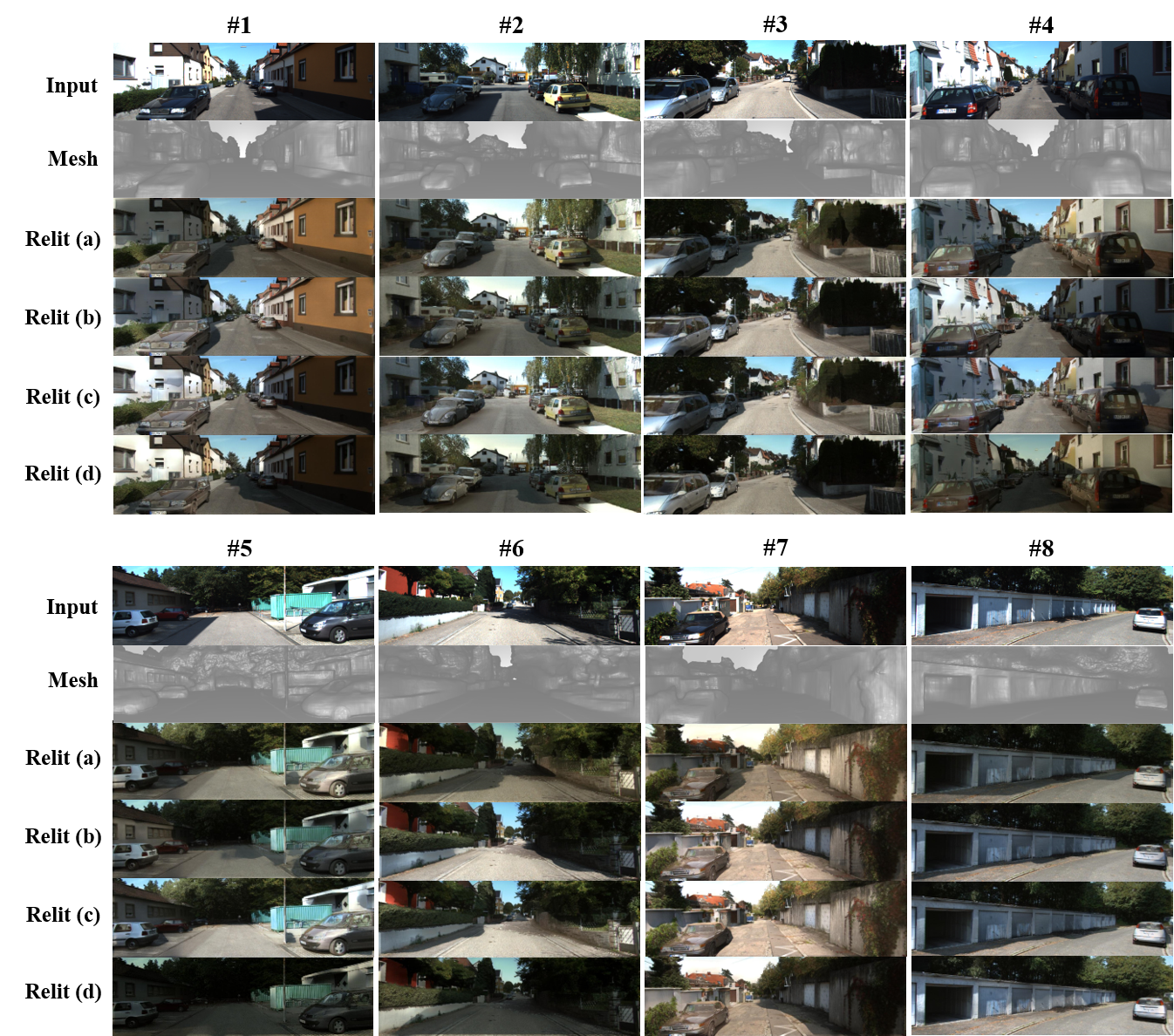}
\caption{\small \textbf{SIMBAR On KITTI:} SIMBAR relights KITTI road driving scenes. Using the \#1 column an example, SIMBAR takes the single frame as input, then creates the 3D scene mesh with geometry priors, and finally generates 4 relit results as shown in Relit (a) (b) (c) (d), with different shadow and lighting conditions. Note that, KITTI scenes are collected around noon time with strong contrasted shadow cast on the ground, which are difficult for the relighting networks to remove. As shown in column \#2 relit results, there are some shadow residuals left. Sometimes, the textures and colors are slightly washed out from the original during relighting (refer column \#7 relit results). Overall, thanks to the depth backbone improvements using dense vision transformer and fine tuned on KTTI dataset, the meshes here have fairly significant complexity and detail leading to robust shadow generation. 
\label{fig:simbar_kitti_supp}}
\end{figure*}

\vspace{-.05in}
 In this section, we present more results on datasets including DIV2K (Fig.~\ref{fig:simbar_div2k_supp}), BDD100K (Fig.~\ref{fig:simbar_bdd100k_supp}), and KITTI Object 2D (Fig.~\ref{fig:simbar_kitti_supp}). These are provided with individual frames, thus the MVR method cannot be applied to datasets prepared in this manner.

% \section{Method Comparison Summary}
% We present the following aspects of MVR-I and SIMBAR methods with advantages and disadvantages of different approaches: 
% \begin{enumerate}[noitemsep]
% \item \textbf{Application:} MVR method is restrained in scenes collected in multiview manner. SIMBAR works on a multitude of real-world datasets. From an application perspective, MVR requires multiview-camera setup for ideal data curation. 
% SIMBAR takes a single input frame, and  generalizes across different real-world datasets.
% \item \textbf{Mesh Reconstruction:} COLMAP fails to converge when lacking feature matching between frames, dynamic moving objects are missing in the scene. For SIMBAR, no feature matching required, but using a single-frame based mesh reconstruction. The generated scene mesh contains dynamic objects.
% \item \textbf{Compute Cost:} MVR method is compute heavy and time consuming, because it constructs the scene mesh for the entire sequence.  And SIMBAR produces a relatively lightweight scene mesh for a single frame. 
% \item \textbf{Video Sequence Relit Consistency:} In addition, since the MVR-Improved method takes a KITTI video sequences,  it is more temporally consistent for or relit result. For SIMBAR, there are relatively more variance.
% \item \textbf{Objects Outside FOV:} For objects outside FOV, MVR-I can generate shadow for these types of objects, as it has access to multiview information. However, for SIMBR, with a single image, it cannot relight objects outside of FOV.
% \end{enumerate}

% \end{document}

\end{document}